\documentclass[3p,twocolumn]{elsarticle}

\usepackage{hyperref}

\usepackage{booktabs,threeparttable}
\usepackage{tabularx} % in the preamble
\usepackage{array,multirow}
\usepackage{graphicx}
\DeclareGraphicsExtensions{.pdf,.png,.jpg,.eps}

\usepackage{subcaption}

\usepackage{amsmath}
\usepackage{pifont}% http://ctan.org/pkg/pifont
\usepackage{amsfonts}
\usepackage{mathtools}
\usepackage{xcolor}
\usepackage{blindtext}
\usepackage{contour}
\usepackage[ruled,linesnumbered]{algorithm2e}
\usepackage{algpseudocode}
\SetKwComment{Comment}{$\triangleright$\ }{}
\SetSideCommentRight

\usepackage{standalone}
\usepackage{pgfplots}

\newcommand{\pasr}{\mathcal{A}}
\newcommand{\pssr}{\mathcal{S}}
\newcommand{\evrn}{\mathcal{R}}
\newcommand{\func}{\mathcal{F}}
\newcommand{\pVs}{\varepsilon}  % psuedo Vs

\newcommand{\Lf}{\pmb{L}}
\newcommand{\Vol}{\pmb{V}}

\newcommand{\Vs}{\boldsymbol{\theta}}
\newcommand{\Xy}{\mathfrak{\textbf{x}}}

\newcommand{\Scale}{\contour[1]{black}{$\zeta$}}

\newcommand{\Stimes}{{\mkern-2mu\times\mkern-2mu}}

\newcommand{\cmark}{{\ding{51}}}
\newcommand{\xmark}{{\ding{55}}}
   
\newcommand{\uds}{\texttt{\char`_}}

\graphicspath{{./graphics/}}

\begin{document}

\begin{frontmatter}

  \title{3DVSR: 3D EPI Volume-based Approach for Angular and Spatial Light field
    Image Super-resolution}

  \author[addrUni]{Trung-Hieu Tran\corref{mycorrespondingauthor}}
  \cortext[mycorrespondingauthor]{Corresponding author}

  \author[addrUni]{Jan Berberich}
  \author[addrUni]{Sven Simon}

  \address[addrUni]{Institute of Parallel and Distributed Systems, Univeristy of
    Stuttgart, 70569 Stuttgart, Germany}

  \begin{abstract}
    Light field (LF) imaging, which captures both spatial and angular information of
    a scene, is undoubtedly beneficial to numerous applications. Although various
    techniques have been proposed for LF acquisition, achieving both angularly and
    spatially high-resolution LF remains a technology challenge.
    In this paper, a learning-based approach applied to 3D epipolar image (EPI) is
    proposed to reconstruct high-resolution LF.
    Through a 2-stage super-resolution framework, the proposed approach effectively
    addresses various LF super-resolution (SR) problems, i.e., spatial SR, angular
    SR, and angular-spatial SR.
    While the first stage provides flexible options to up-sample EPI volume to the
    desired resolution, the second stage, which consists of a novel EPI volume-based
    refinement network (EVRN), substantially enhances the quality of the
    high-resolution EPI volume.
    An extensive evaluation on 90 challenging synthetic and real-world light field
    scenes from 7 published datasets shows that the proposed approach outperforms
    state-of-the-art methods to a large extend for both spatial and angular
    super-resolution problem, i.e., an average peak signal to noise ratio
    improvement of more than 2.0 dB, 1.4 dB, and 3.14 dB in spatial SR $\Stimes 2$,
    spatial SR $\Stimes 4$, and angular SR respectively.
    The reconstructed 4D light field demonstrates a balanced performance
    distribution across all perspective images and presents superior visual quality
    compared to the previous works.
  \end{abstract}


\end{frontmatter}

%\linenumbers

\section{Introduction}
\label{sec:intro}
The main advantage of a light field (LF) image over the conventional image is its
high dimensional data. A LF image contains both directional information
and spatial information instead of only spatial information as in conventional
image. This rich-content property of light field brings a
great benefit to numerous applications~\cite{Wu2017}.
In general,  light field acquisitions can be categorized into three main
classes:  multi-sensor capturing~\cite{Wilburn2005}, time-sequential
capturing~\cite{Unger2003} and multiplexed
imaging~\cite{Adelson1992,Ng2005,Lumsdaine2009}.
The multi-sensor capturing approach requires an array of image sensors
distributed on a planar or spherical surface to simultaneously capture
light field samples from different viewpoints.
The time-sequential capturing approach, on the other side, uses a single image
sensor to capture multiple samples of the light field through multiple exposures.
The typical approach uses a sensor mounted on a mechanical gantry to measure the
light field at different positions~\cite{Unger2003}. The multiplexed
imaging encodes the high dimensional light field into a 2D sensor plane, by
multiplexing the angular domain into the spatial domain. One popular example of
this acquisition approach is plenoptic camera~\cite{Adelson1992, Ng2005,
  Lumsdaine2009}, in which a microlens array is  placed in between a main lens and an image sensor.

Each acquisition method has its own advantages and disadvantages. The multi-sensor
capturing approach is generally more expensive but allows capturing very high
spatial resolution of a dynamic screen.
The time-sequential capturing approach is inexpensive and can capture very high
spatial resolution but suffers from very long capturing time which makes it less
preferable for capturing a dynamic scene. Multiplexed imaging approach is
inexpensive and can handle dynamic scenes but produces low spatial resolution
images. Besides, all acquisition approaches impose a trade-off between
angular resolution and either spatial resolution or temporal resolution, i.e.,
reducing the cost of the sensor by using low-resolution sensors in exchange to
increasing number of sensors for higher angular resolution; increasing capturing
time to capture dense angular samples of the scene; increasing the
number of micro-lenses to increase the spatial resolution but at the
same time reducing the number of angular samples.
The existing challenges in capturing high-resolution light field images limit
its promotion to practical applications and motivates research in light field
super-resolution (SR)~\cite{Cheng2019}.

\begin{figure}[t]
  \centering
  \begin{minipage}[b]{\linewidth}
    \begin{minipage}[b]{.49\linewidth}
      \centering
      \includegraphics[width=\textwidth]{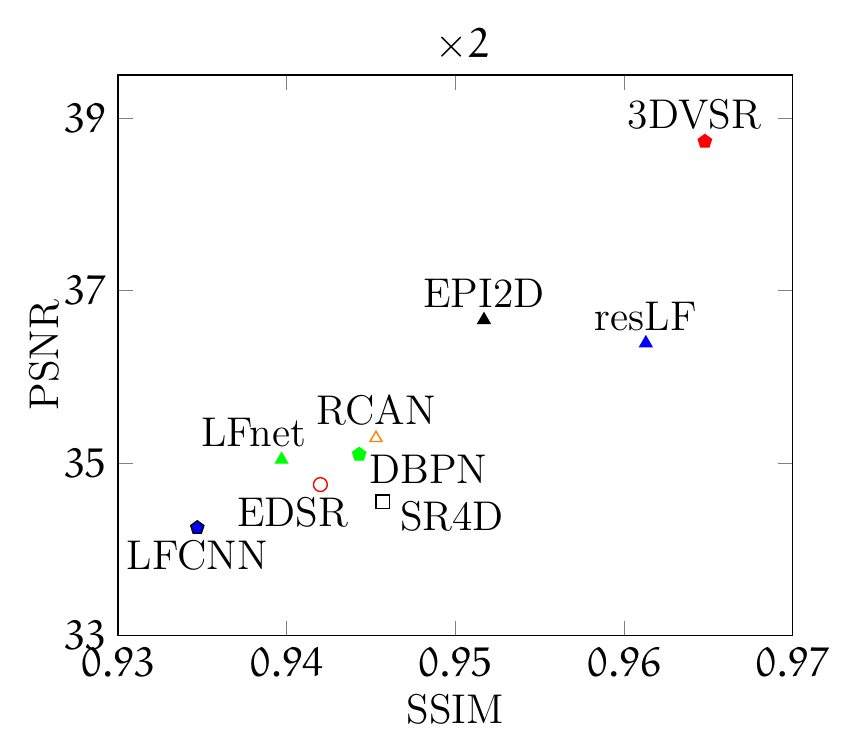}
    \end{minipage}
    \hfill
    \begin{minipage}[b]{0.467\linewidth}
      \centering
      \includegraphics[width=\textwidth]{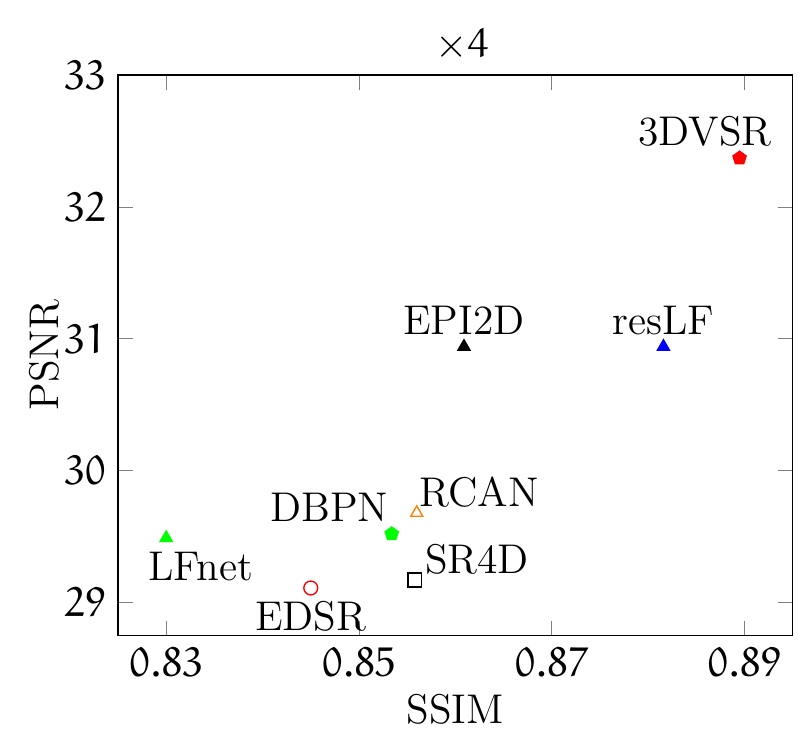}
    \end{minipage}
  \end{minipage}
  \hfill
  \caption{
    Spatial super-resolution results achieved by the proposed approach (3DVSR) and state-of-the-art SSR methods.
    The data points are averages of the performance metrics from 90 test scenes
    of 7 public datasets. (\cite{Wanner2013,
      Honauer2017, Shi2019,StanfordGantry,StanfordLytro, InriaLytro,
      Rerabek2016}).
    The proposed approach outperforms single image SR methods
    (EDSR~\cite{Lim2017}, RCAN~\cite{Zhang2018rcan},
    DBPN~\cite{Haris2018}) and LF SSR methods (pcabm~\cite{Farrugia2017},
    LFCNN~\cite{Yoon2017}, LFnet~\cite{Wang2018}, EPI2D~\cite{Yuan2018}, SR4D~\cite{Yeung2019},
    resLF~\cite{Zhang2019}).
  }
  \label{fig:exp_ssr}
\end{figure}

An LF image can be characterized by a 4D array structure comprised of
two spatial dimensions and two angular dimensions.
In practice, it is considered as a 2D array of sub-aperture images (SAIs)
that captures a scene from different perspectives.
LF super-resolution distinguishes between spatial super-resolution
(SSR), angular super-resolution (ASR), and angular-spatial super-resolution
(ASSR).
SSR aims for increasing the resolution of each SAI while ASR aims for
generating new perspective images.
ASSR, as its name indicates, involves both SSR and ASR and results in higher
resolutions in all dimensions of an LF image.
A straightforward approach for SSR is applying a single image super-resolution
(SISR) technique in which each SAI is up-sampled independently.
Thanks to comprehensive training dataset and mature development in the field of
SISR, state-of-the-art approaches such as VDSR~\cite{Kim2016},
EDSR~\cite{Lim2017}, RCAN~\cite{Zhang2018rcan}, RDN~\cite{Zhang2018a} can
already reconstruct relatively good quality
SAIs~\cite{Cheng2019}.
However, failing to incorporate angular information, this type of approach is
easily surpassed by recent LF SSR methods (EPI2D~\cite{Yuan2018},
ResLF~\cite{Zhang2019}, SR4D~\cite{Yeung2019}).
Compared to SSR, there is less research attention to ASR and ASSR.
Existing approaches such as VSYN~\cite{Kalantari2016}, LFCNN~\cite{Yoon2017},
LFSR~\cite{Gul2018}, Wang et al.~\cite{Wang2018a}, and Wu et al.~\cite{Wu2019,
  Wu2019a} employ convolutional neural network (CNN) for predicting novel SAI.
However, their performances are still limited.

In this paper, a 3D EPI volume-based super-resolution approach (3DVSR) is
proposed for dealing with LF super-resolution problems, i.e., ASR, SSR, and
ASSR. As opposed to existing methods that rely on 2D slices of LF, i.e.,
epipolar images (EPIs)~\cite{Yuan2018} or SAIs~\cite{Zhang2019, Wang2018}, our
approach makes use of a 3D projected version of LF, also known as 3D EPI
volume~\cite{Heber2017}. The advantage of this volume structure is incorporating
both spatial information in SAI and angular information in EPI. In addition, to
fully exploit this 3D structure, we proposed a novel EPI volume refinement
network (EVRN) built on 3D convolution operations and efficient deep learning
techniques, i.e., global/local residual learning, dense connection, multi-path
learning, and attention-based scaling. Experimental results show that EVRN
substantially improves the reconstruction quality in all LF super-resolution
problems. Fig.~\ref{fig:exp_ssr} presents the performance of the proposed
approach compared to state-of-the-art methods for SSR. Peak signal to noise
ratio (PSNR) and structural similarity index measure (SSIM) are employed as
performance metrics. The proposed approach outperforms the existing approaches
to a large extent in both SSR $\Stimes 2$  and $\Stimes 4$. Detail comparisons
and evaluations are further discussed in Section~\ref{sec:exp}.

The main contributions of this work are as follows. First, we proposed a novel
3D EPI volume-based framework for addressing various LF SR problems, i.e., ASR,
SSR, and ASSR. Specifically, the framework comprises two consecutive stages,
i.e., preliminary up-sampling stage and volume-based enhancement stage. The first
stage allows different options to be used for up-sampling the input volume to
the desired resolution. Then, in the second stage, the novel EVRN corrects the
high-frequency information by incorporating both spatial information in SAI and
angular information in 2D EPI to reconstruct a high-quality LF image.
Such a two-stage model was first introduced by Fan et al.~\cite{Fan2017} for
reconstructing a high-resolution SAI, but our approach is fundamentally
different. In~\cite{Fan2017}, outputs of the first stage are subjected to a view
registration before feeding to a CNN to improve the quality of a targeted view.
Although this method generates more references to support the enhancement of the
targeted view, it destroys the EPI structure. Our approach, in contrast,
preserves this structure in an EPI volume and introduces a novel EVRN to exploit
this information for enhancing multiple views simultaneously. Compared to
previous works which employed residual, dense, and attention techniques to
reconstruct a high-resolution 2D image (RDN~\cite{Zhang2018a},
RCAN~\cite{Zhang2018rcan}), the novelty of our EVRN lies in the introduction of
attention-based multi-path learning, which targets spatial and angular aspects
of the feature maps. As discussed in Section~\ref{sec:exp}, this technique
allows us to improve the performance of EVRN and, therefore, leads to a
substantial enhancement of EPI volume output from the preliminary up-sampling
stage.
Secondly, a simple but effective Deep CNN model is proposed for preliminary
up-sampling angular dimensions. Although its output quality is later greatly
improved by EVRN, this model itself already surpasses existing approaches in
ASR. Thirdly, an extensive evaluation is conducted on 90 challenging synthetic
and real-world LF scenes from 7 public LF datasets. In this evaluation, we
analyzed the performance of the proposed approach and compared it to
state-of-the-art approaches.

The remainder of the paper is organized as follows.
Section~\ref{sec:related_work} briefly discusses related works categorized into
optimization-based approaches and learning-based approaches. An overview of LF
representation along with important notations is presented in
Section~\ref{sec:notation}. Section~\ref{sec:method} and Section~\ref{sec:exp}
respectively discuss the proposed approach and experimental results. Finally, we
draw a conclusion in Section~\ref{sec:conclusion}.

\section{Related Work}
\label{sec:related_work}
In general, previous works can be categorized into two groups:
optimization-based approaches and learning-based approaches.

\subsection{Optimization-based approaches}
In this type of approach, LF super-resolution is formulated as an optimization
problem which typically consists of a data fidelity term directly composed from
input LF image and a regularization term based on known priors.
There are two main types of data terms that are used in the literature.
One of them penalizes the coherence between low and high-resolution LF
image pairs~\cite{Mitra2012, Bishop2012, Alain2018, Rossi2018}.
The other enforces the Lambertian consistency across the directional dimension
by warping SAIs from different view angles using pre-computed disparity maps
~\cite{Wanner2014, Tran2018, Rossi2018}.
Compared to the data fidelity term, the choices of regularization terms are more
diverse. Each work proposed to use a different prior in order to achieve a
better output quality and with a feasible computation effort, i.e., total
variation (TV) \cite{Wanner2014}, bilateral TV\cite{Tran2018}, Markov Random
Field (MRF) \cite{Bishop2012}, Gaussian \cite{Mitra2012}, Graph-based \cite{Rossi2018a},
sparsity \cite{Alain2018}.

In \cite{Bishop2012}, Bishop et al. studied an explicit image formation model
that characterizes the light field imaging process by spatially-variant point
spread functions (PSFs). The PSFs were derived under Gaussian optics
assumptions and employed in a Bayesian framework for super-resolution.
In \cite{Mitra2012}, Mitra et al. showed that 4-D patches of different
disparities have different intrinsic dimensions and proposed to learn a Gaussian
prior for each quantized disparity value. These priors were then employed to
inference high-resolution 4-D patches under the Maximum a posterior (MAP)
criterion.
LF super-resolution was modeled as a continuous optimization problem using a variational
framework in \cite{Wanner2014}.
Disparity maps were extracted by local estimation of pixel-wise slope in
EPI.
The data fidelity term was constructed by warping surrounding views with the estimated
disparity maps and masking with occlusion maps, while total variation was used for
regularization.
In \cite{Tran2018}, Tran et al. treated LF super-resolution as a multi-frame
super-resolution problem in which degradation process is modeled by three
operators: down-sampling, blurring, and warping.
A variational optimization approach was employed to estimate disparity maps used by
the warping operator and  bilateral TV was employed as an image prior.
In \cite{Alain2018}, Alain et al. proposed a patch-based super-resolution approach
making use of a 5D transform filter that consists of 2D DCT transform, 2D
shape-adaptive DCT and 1D haar wavelet.
By a proper selection of 5D patches, a transformed signal exposes a high degree of
sparsity which can be employed as a prior to regularise a $L^2$ data term.
In \cite{Rossi2018a}, Rossi et al. proposed an approach which couples two
data-terms with a graph-based regularizer.
A graph-based prior regularizes high-resolution SAIs by enforcing the geometric light
field structure.
Block matching was employed in their work for the estimation of disparity values
and the construction of the graph map.

\subsection{Learning-based approaches}
\begin{figure}[t]
  \centering
  \begin{minipage}[b]{.8\linewidth}
    \begin{minipage}[b]{.34\linewidth}
      \centering
      \includegraphics[width=\textwidth]{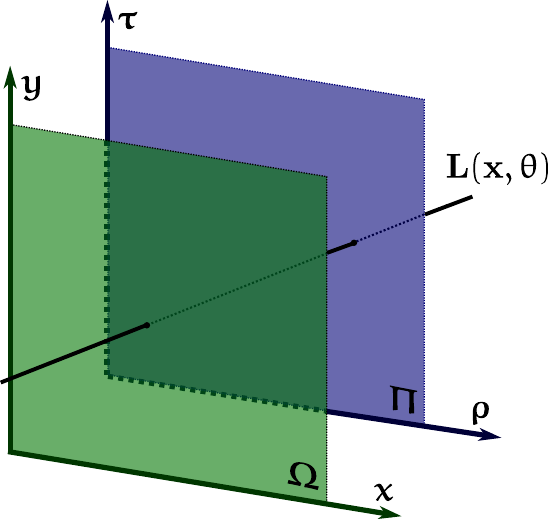}
      \centerline{(a)}\medskip
    \end{minipage}
    \hfill
    \begin{minipage}[b]{0.3\linewidth}
      \centering
      \includegraphics[width=\textwidth]{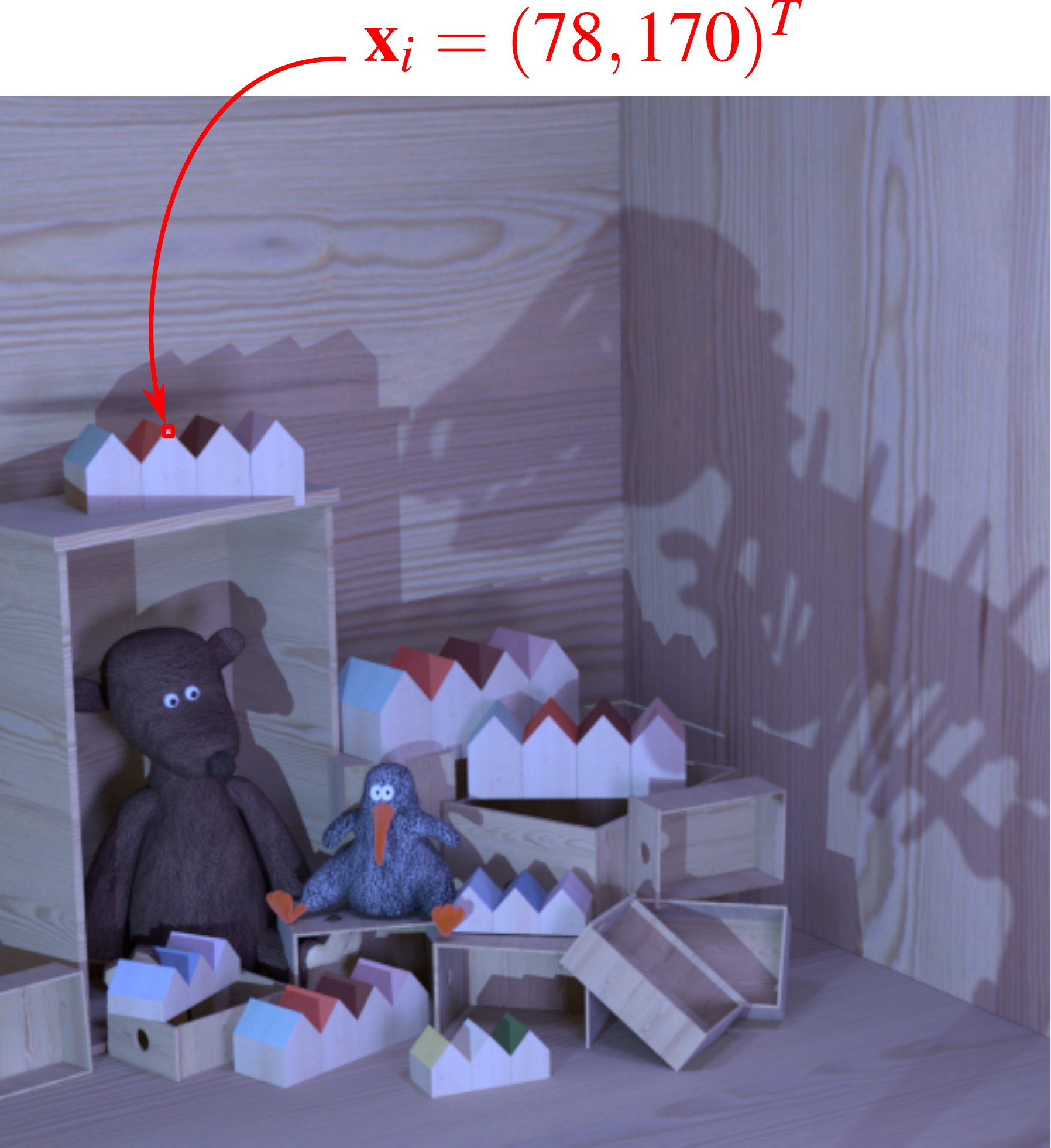}
      \centerline{(b)}\medskip
    \end{minipage}
    \hfill
    \begin{minipage}[b]{0.3\linewidth}
      \centering
      \includegraphics[width=\textwidth]{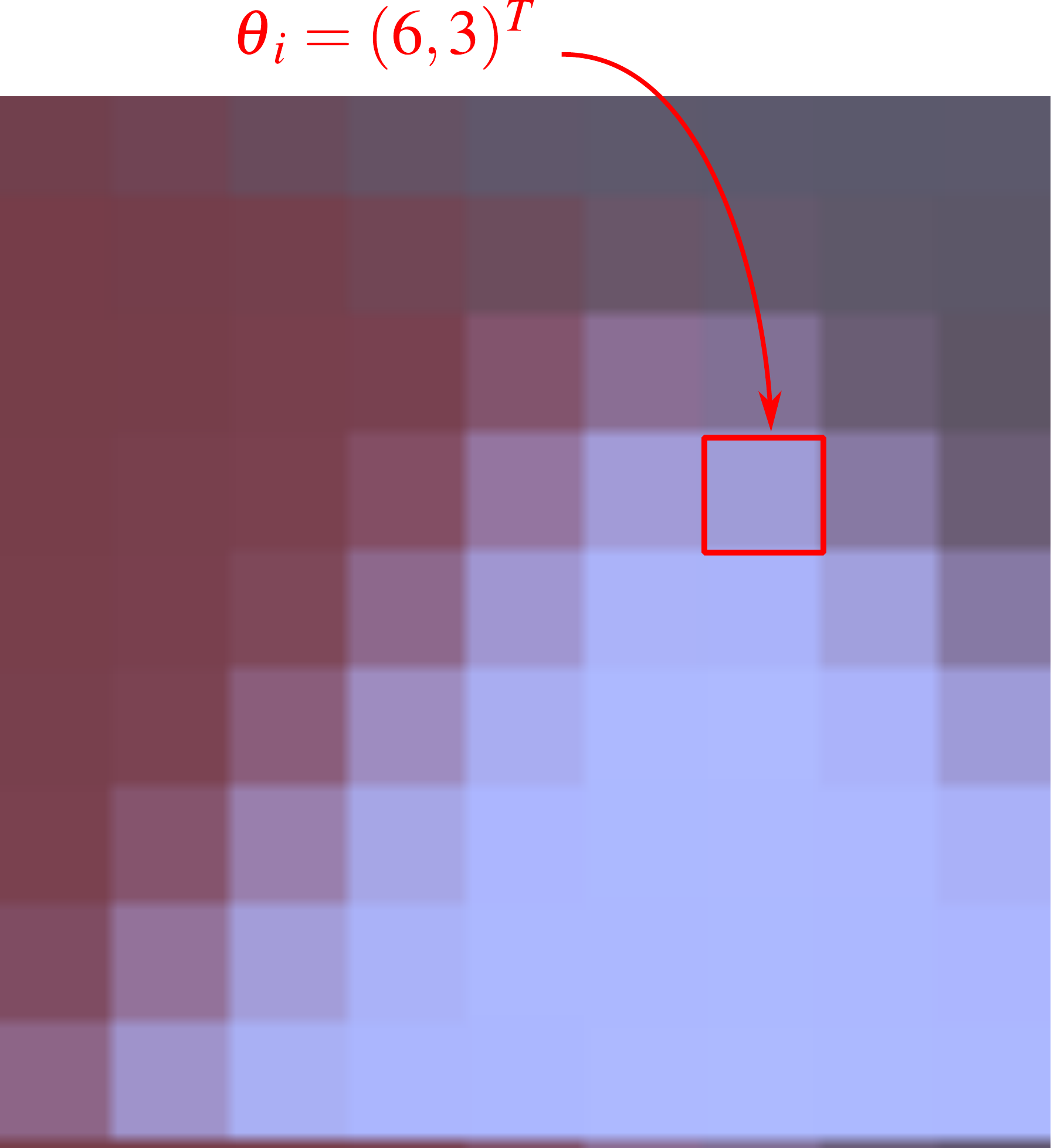}
      \centerline{(c)}\medskip
    \end{minipage}
  \end{minipage}
  \caption{4D Light field representation. (a) Two-plane
    parameterisation. (b) A sub-aperture image of scene \textit{dino}~\cite{Honauer2017} at angular location $\Vs_i=(6,3)^T$.
    (c) An angular patch image at spatial location $\Xy_i=(78,170)^T$.}
  \label{fig:two_plane}
\end{figure}

The earliest work which applies deep learning for reconstructing high-resolution LF is~\cite{Yoon2015}. The authors employed the SISR approach from~\cite{Dong2014} for SSR and proposed a CNN-based solution for ASR in which novel views depending on its position will be synthesized from a vertical pair, a horizontal pair, or four neighbors. An improved version was described in~\cite{Yoon2017} where SISR was applied to each SAI separately, and learning variables were shared in the ASR network.
In~\cite{Kalantari2016}, Kalantari et al. proposed to generate novel SAIs by exploiting disparity information in a two-stage CNN. The first stage predicts disparity maps from a pre-computed cost-volume, and the second stage synthesizes novel views from input images that are warped using predicted disparity maps.
In~\cite{Farrugia2017}, a patch-based approach that employs linear subspace projection was presented. A linear mapping function between low and high subspaces of low and high LF patch volume was learned from a training dataset and applied to new LF images. The authors used block matching to find matched 2D patches and extract aligned patch volumes. Principal component analysis (PCA) was then employed to reduce patch volumes' dimensions and project them into subspaces. The mapping function was computed in the form of a $l_2$-norm regularized least square problem.
Fan et al.~\cite{Fan2017} proposed a two-stage approach for SSR. In the first stage, each SAI was upscaled using the SISR method from VDSR~\cite{Kim2016}. The output was then registered to a reference SAI by locally searching similar patches. Both reference image and registered images were fed into a CNN network in the second stage to reconstruct high-resolution SAI at the reference position.
Similarly, Yuan et al.~\cite{Yuan2018} proposed another two-stage approach. In the first stage, EDSR~\cite{Lim2017} was employed to super-resolute SAIs. In the second stage, the output of SISR was then enhanced by a refinement CNN which relies on 2D EPI.
In~\cite{Gul2018}, Gul et al. proposed an approach targeting lenslet images captured by plenoptic cameras~\cite{Adelson1992,Ng2005}. Microlens image patches were used as input to two separate networks, i.e., one for ASR and the other for SSR. However, the fully connected layers employed in these networks limited its application to a certain angular resolution.
Wang et al.~\cite{Wang2018} developed a bidirectional recurrent CNN approach for spatially up-sampling 4D LF images. They employed multi-scale fusion layers for future extraction to accumulate contextual information. Two networks for vertical and horizontal image stacks were learned separately, and a stack generalization technique was employed to obtain a complete set of images.
In~\cite{Wang2018a}, the authors proposed a two-step approach to synthesize novel views. In the first step, a learnable interpolation approach is employed to generate intermediate volumes, which are then refined in the second step through a 3D CNN. However, since the architecture of the 3D CNN is simple, the performance of this refinement step is limited.
In~\cite{Wu2019} and ~\cite{Wu2019a} Wu et al. proposed two novel approaches for LF view synthesis based on EPI structure. In~\cite{Wu2019}, a set of shared EPIs is created for a discrete set of shear values which is then up-sampled and scored by an evaluation CNN to create a cost volume. The 2D slices output from fusing the cost volume are then subjected for a high-resolution EPI calculation based on a pyramid decomposition-reconstruction technique. In \cite{Wu2019a}, the authors model the view synthesis as a learning-based detail restoration on 2D EPIs. They proposed a three-step framework, namely ``blur-restoration-deblur'' that employs the blurring operator for blur step, a CNN for restoration step, and a non-bind deblur operation for the last step.
To fully exploit the 4D structure of LF images, Yeung et al.~\cite{Yeung2019} proposed to apply 4D convolution for SSR. The 4D convolution function was implemented as spatial-angular separable convolution, which allows extracting feature maps from both spatial and angular domains.
In~\cite{Zhang2019}, Zhang et al. proposed a residual CNN-based approach for reconstructing LF images with higher spatial resolution. The network takes in image stacks from four different angles and predicts a high-resolution image at the center position. According to the difference in angular position, it requires six different networks for a complete reconstruction of the 4D LF image.
\begin{figure}
  \centering
  \begin{minipage}[b]{\linewidth}
    \includegraphics[width=0.8\linewidth]{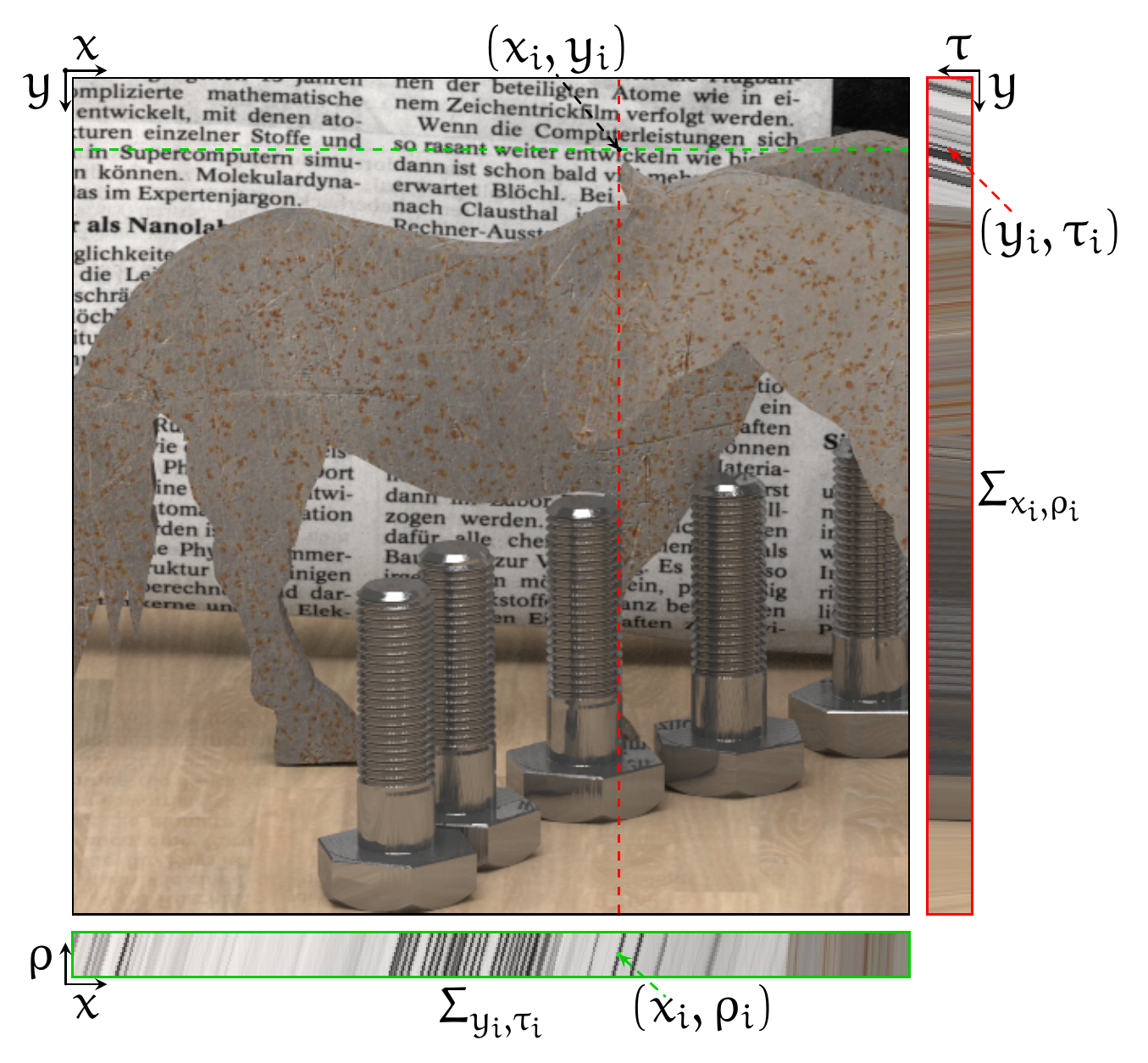}
  \end{minipage}
  \caption{2D EPI projection of 4D LF scene \textit{horses}~\cite{Wanner2013}.
    {\it top-left} SAI at $\Vs_i=[\rho_i,\tau_i]^T$; {\it top-right} vertical EPI ($\Sigma_{x_i,\rho_i}$);
    {\it bottom} horizontal EPI ($\Sigma_{y_i, \tau_i}$).
  }
  \label{fig:epi}
\end{figure}

\section{Light field Notation}
\label{sec:notation}
Light field refers to the concept of acquiring a complete description of light
rays emitted from a scene and traverse in space. It can be well parameterized by
the 7D plenoptic function
$\Lf\big(p_x,p_y,p_z,\theta,\phi,\lambda,t\big)$~\cite{Adelson1992} which
returns the radiance (intensity) of a line beam observed at a point
$\big(p_x,p_y,p_z\big)$ in space, along a grazing direction
$\big(\theta,\phi\big)$, at a time $t$ and wavelength $\lambda$.
In practice, it is of interest to capture only a snapshot of the function (at
a fixed time) and simplify the spectral information by using only 3 color
components (i.e. red, green, blue). By removing the time and wavelength parameters, we
are left with a 5-parameters function, also known as 5D plenoptic function
\cite{Levoy1996}, which allows us
to describe the intensity of any light ray in 3D space regardless the viewpoint
position and the change of light direction.
One more parameter can be omitted from the function, if one assumes that the
direction of a light ray is unchanged and considers only the light intensity
being visible at a specific position, i.e. placing of cameras.
This is also the common setup of light field imaging which results in a 4D
light field~\cite{Levoy1996} or Lumigraph~\cite{Gortler1996}.
Under the two plane parameterization, each ray is indexed
with a 4D coordinate by its intersection with two parallel planes, as depicted
in Figure~\ref{fig:two_plane}a.

\begin{equation}
  \Lf\colon \Omega \Stimes \Pi \rightarrow \mathbb{R},\qquad  (\Xy,\Vs) \rightarrow
  \Lf(\Xy,\Vs)
\end{equation}
where $\Xy =[x,y]^T$ and $\Vs = [\rho,\tau]^T$ denote coordinate pairs in the spatial
plane $\Omega \subset \mathbb{R}^2$ and in the directional plane $\Pi \subset
\mathbb{R}^2$ respectively.
In practice, the value of this function can be a vector of 3 color components
(i.e., RGB color light field) and the two planes are discretized by the
sampling rate of capturing devices (e.g., sensor size, number of cameras,...).

For a better observation and analysis, various ways to visualize 4D LF are
introduced in the literature.
Common visualizations are using 2D slices, i.e., sub-aperture image (SAI), angular
patch image (API), or epipolar image (EPI). By fixing the directional index $\Vs=\Vs_i$
or the spatial index $\Xy=\Xy_i$, one can respectively obtain a SAI
$\Lf(\Xy,\Vs_i)$ or an API $\Lf(\Xy_i,\Vs)$.
Fig.~\ref{fig:two_plane} (b),(c) are examples of a sub-aperture image and an
angular patch extracted from a 4D synthetic light field image,
\textit{dino}~\cite{Honauer2017}.
The resolution of this light field is 512$\Stimes$512$\Stimes$9$\Stimes$9, where
512$\Stimes$512 is the spatial resolution and 9$\Stimes$9 is the angular resolution.
The angular patch consists of pixels, each at the same spatial location
$\Xy_i=(78,170)^T$ on a sub-aperture.
An EPI can be acquired by fixing one index in the spatial plane and one index in
the angular plane while varying the remaining two indices.
By fixing vertical indices, i.e., $y\!=\!y_i, \tau\!=\!\tau_i$, we have the horizontal
EPI \(\Sigma_{y_i,\tau_i}(x,\rho)=\Lf\big([x,y_i]^T,[\rho,\tau_i)]^T\big)\).
A similar procedure applies to vertical EPI \(\Sigma_{x_i,\rho_i}(y,\tau)\).
Fig.~\ref{fig:epi} illustrates the two types of EPIs.

As pointed out in~\cite{Wanner2014}, a point in 3D space is projected onto
a line in EPI whose slope is decided by the depth of this point.
For example, in Figure~\ref{fig:epi} the point located at $(x_i,y_i)$ projected
onto two lines in horizontal EPI
($\Sigma_{y_i,\tau_i}$) and vertical EPI ($\Sigma_{x_i,\rho_i}$).
Notice that all letters crossed by the line $y=y_i$ (i.e., dotted green line)
reside on the same depth
and result in identical slope in horizontal EPI $\Sigma_{y_i,\tau_i}$.
This property of EPI represents the geometric information of the scene and was
exploited in many LF image processing applications (i.e. disparity
estimation~\cite{Heber2017, Wanner2014}, super-resolution~\cite{Wanner2014,
  Yuan2018}).
In this work, we consider a 3D version of EPI, namely EPI
volume~\cite{Heber2017}, in which the second spatial dimension is added.
The intuition behind this volume data is the combination of both spatial
coherence (i.e. $x$ vs. $y$) and EPI coherence (i.e. $x$ vs. $\rho$)
that can be employed for reconstructing high-quality LF image.
Horizontal EPI volume $\Vol_{\tau_i}$ and vertical EPI volume $\Vol_{\rho_i}$
are defined in Equation~\ref{eq:3depi}.
An EPI volume is an orthogonal 3D slide through 4D LF and can be considered as a
stack of 2D EPI along a spatial axis, e.g., $\Vol_{\tau_i}$ is constructed by
stacking horizontal EPI $\Sigma_{y_i,\tau_i}$ along spatially vertical axis $y$.
\begin{equation}
  \begin{split}
    \Vol_{\tau_i}\colon \mathbb{R}^3 \rightarrow \mathbb{R},\qquad
    (x,\rho, y) \rightarrow
    \Lf\left(\begin{bmatrix}x\\ y\end{bmatrix},\begin{bmatrix}\rho\\ \tau_i\end{bmatrix}\right) \\
    \Vol_{\rho_i}\colon \mathbb{R}^3 \rightarrow \mathbb{R},\qquad
    (y,\tau, x) \rightarrow
    \Lf\left(\begin{bmatrix}x\\ y\end{bmatrix},\begin{bmatrix}\rho_i\\ \tau\end{bmatrix}\right)
  \end{split}
  \label{eq:3depi}
\end{equation}

\begin{figure}[t]
  \centering
  \begin{minipage}{0.8\linewidth}
    \centering
    \includegraphics[width=\linewidth]{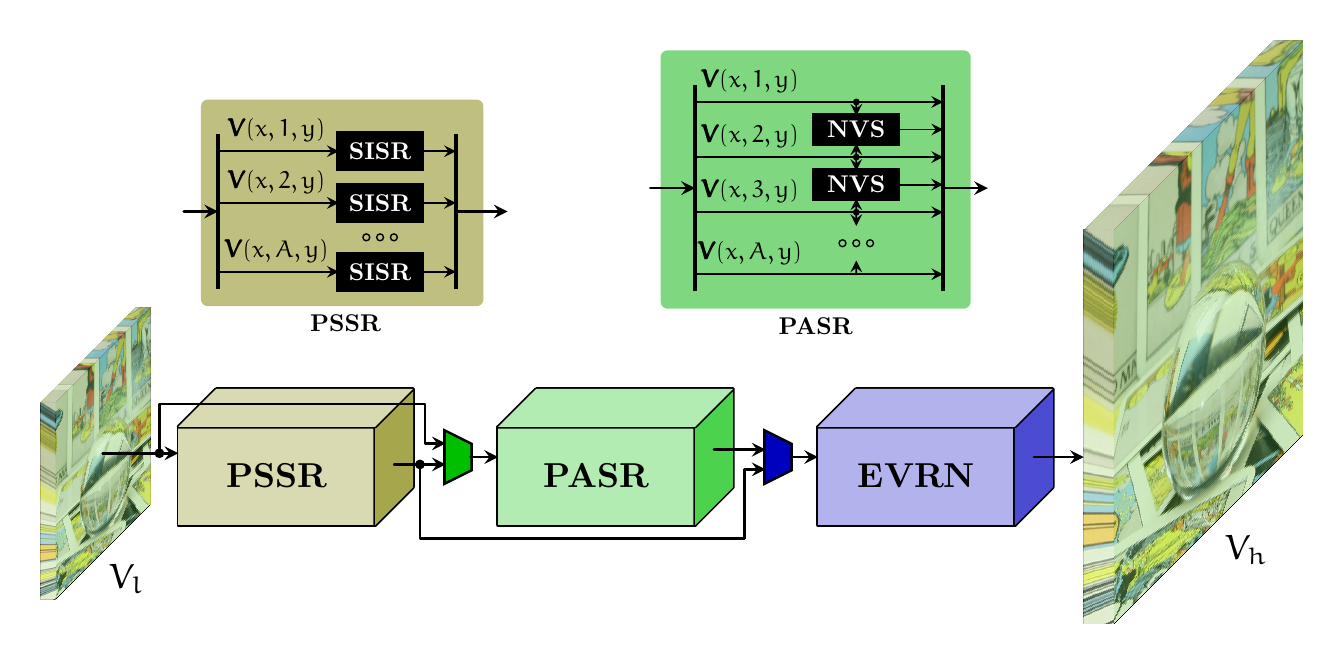}
  \end{minipage}
  \caption{Overview of 3D EPI volume SR framework. The framework consists of a
    preliminary up-sampling stage and an enhancement stage. The earlier stage includes
    a preliminary spatial SR block (PSSR) and a preliminary angular
    SR block (PASR). The later stage includes an EPI volume refinement network (EVRN).}
  \label{fig:overall}
\end{figure}
This work assumes that 4D LF image has the resolution of $W\Stimes H \Stimes
A \Stimes A$, where $H, W \in \mathbb{N}^{+}$ respectively represent the height
and width of each SAI and $A\!=\!2K+1, K\!\in\!\mathbb{N}^{+}$ denotes the angular
resolution.
The resolutions of $\rho$-axis and $\tau$-axis are set equally, since this
square array of views is commonly used in
literature~\cite{Farrugia2017,Yoon2017,Wang2018,Yuan2018,Yeung2019,Zhang2019} and available
in public dataset~\cite{  Wanner2013,Honauer2017,Shi2019,StanfordGantry,
  StanfordLytro,InriaLytro,Rerabek2016}.
The resolution of horizontal EPI volume
$\Vol_{\tau_i}$ and vertical EPI volume
$\Vol_{\rho_i}$ are then $W\Stimes A \Stimes H$ and $H\Stimes
A\Stimes W$ respectively.

\begin{figure*}[t]
  \centering
  \includegraphics[width=0.95\linewidth]{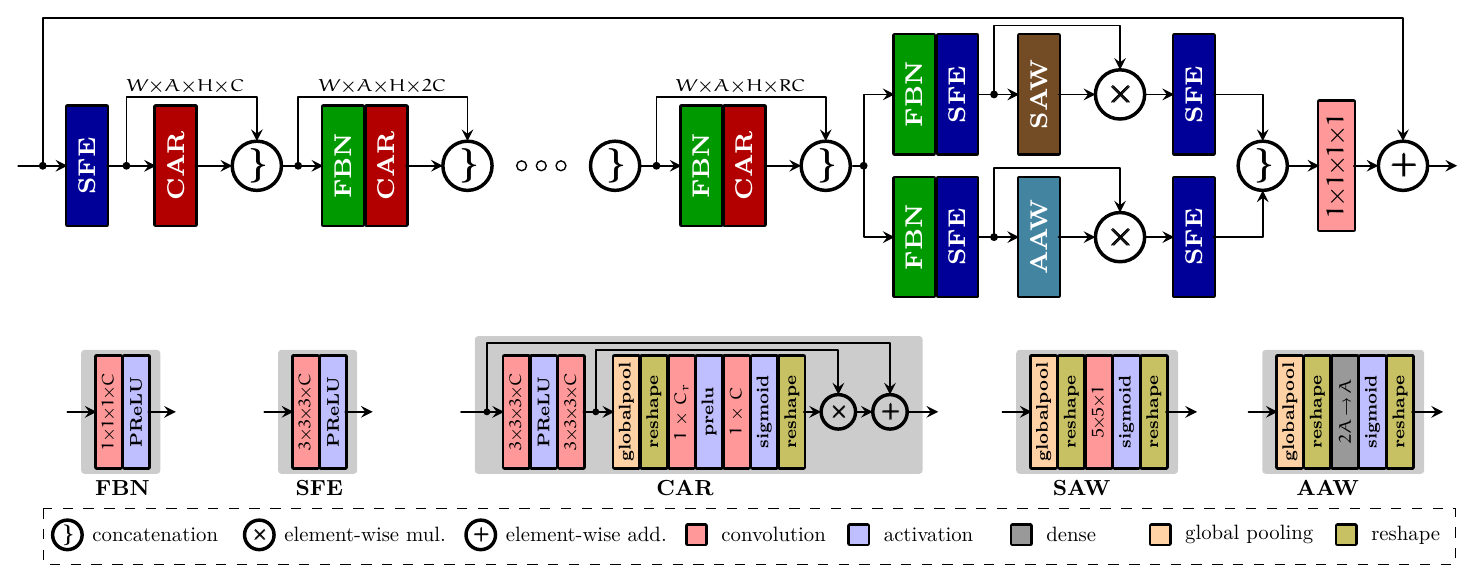}
  \caption{The network architecture of the proposed EPI volume refinement network (EVRN)}
  \label{fig:evrn}
\end{figure*}

\section{Methodology}
\label{sec:method}
\subsection{Overview of the Proposed Approach}
The proposed approach targets the super-resolution of 3D projected versions of a
4D LF image.
Instead of directly up-sampling a 4D LF image (e.g, SR4D~\cite{Yeung2019}), we first
decompose it into a complete set of 3D EPI volumes.
Each EPI volume is then super-resolved to its desired resolution.
The high-resolution sets of EPI volumes are finally merged to form the final
high-resolution 4D output.
Fig.~\ref{fig:overall} illustrates the overall process of the proposed EPI volume
SR framework.
The framework is comprised of two main processing stages: a preliminary up-sampling
stage and a volume enhancement stage. The first stage involves two
tasks: preliminary spatial super-resolution (PSSR) and preliminary angular
super-resolution (PASR).
The second stage includes an EPI volume-based refinement network (EVRN).
Given a 3D EPI volume (\(\Vol_l\)) with the resolution of \(W\Stimes A\Stimes
H\), PSSR and PASR sequentially up-sample the spatial
resolution to \(\Scale_{\Xy} W\Stimes \Scale_{\Xy} H\) and the angular resolution to
\(\Scale_{\Vs} A\), where $\Scale_{\Xy}$ and $\Scale_{\Vs}$ are spatial and
angular scaling factor respectively.
This preliminarily up-sampling volume is provided as an input to EVRN which
will, in turn, refine the 3D EPI structure and return an enhanced
high-resolution volume.
PSSR and PASR can be applied separately or jointly depending on SR applications
(i.e., SSR, ASR, and ASSR).
\begin{algorithm}
  \SetAlgoLined
  \DontPrintSemicolon
  \SetKwFunction{FMain}{VSR}
  \SetKwProg{Pn}{Function}{:}{end}

  \Pn{\FMain{$\Lf_l$, $\pVs$, $\func$}}{
    $\{\Vol\}_{\pVs} := \texttt{Slice}(\Lf_l,\pVs)$
    \Comment*[r]{\small Extract $\pVs$-axis volumes}
    $\{\Vol\}_* := \{\}$
    \Comment*[r]{\small Initialize empty volume set}
    \For{$\Vol$ in $\{\Vol\}_{\pVs}$}{
      $\{\Vol\}_* := \{\Vol\}_* + \func(\Vol)$
    }
    $\Lf_h := \texttt{Merge}(\{\Vol\}_*, \pVs)$
    \Comment*[r]{\small Reconstruct 4D LF}
    \KwRet $\Lf_h$\;
  }
  \caption{Volume-based super-resolution (VSR) function}
  \label{alg:vsr}
\end{algorithm}

\begin{algorithm}
  \SetAlgoLined
  \DontPrintSemicolon
  \KwIn{$\Lf_l$}
  \KwOut{$\Lf_h$}
  $\Lf :=\Lf_l$\;
  \uIf{spatial super-resolution}{
    $\func :=\evrn\big(\pssr(\cdot)\big)$
    \Comment*[r]{\small Apply PSSR and EVRN}
    $\Lf := 0.5\Big(\texttt{VSR}(\Lf,\tau,\func) + \texttt{VSR}(\Lf,\rho,\func)\Big)$\;
  }
  \ElseIf{angular super-resolution}{
    \If{spatial super-resolution}{
      $\func :=\pssr(\cdot)$
      \Comment*[r]{\small Apply only PSSR}
      $\Lf := \texttt{VSR}(\Lf,\tau,\func)$
    }
    $\func :=\evrn\big(\pasr(\cdot)\big)$
    \Comment*[r]{\small Apply PASR and EVRN}
    $\Lf := \texttt{VSR}(\Lf,\tau,\func)$
    \Comment*[r]{\small Upscale $\rho$-axis}
    $\Lf := \texttt{VSR}(\Lf,\rho,\func)$
    \Comment*[r]{\small Upscale $\tau$-axis}
  }
  $\Lf_h:=\Lf$\;
  \caption{Spatial-angular super-resolution of 4D LF}
  \label{alg:assr}
\end{algorithm}

The function $\texttt{VSR}$ in Algorithm~\ref{alg:vsr} describes the application
of the proposed framework for the reconstruction of high resolution LF image.
$\texttt{VSR}$ takes in three paramters: a low-resolution LF ($\Lf_l$), a directional axis
($\pVs\!\in\!\{\rho, \tau\}$), and a volume-based SR function ($\func$).
The directional axis is needed to determine the direction of 3D projection which
will results in either vertical volume ($\pVs\!=\!\rho$) or horizonal volume
($\pVs\!=\!\tau$).
The volume-based SR function encodes the configuration of PSSR and PASR as in
Fig.~\ref{fig:overall}, i.e., only PSSR, only PASR, and both.
The procedure of $\texttt{VSR}$ is as follow.
First, a set of EPI volumes are extracted from the input $\Lf_l$ (line $2$).
Depending on $\pVs$, function $\texttt{Slice}(\cdot)$ will return either
horizontal volume or vertical volume set $\big(\{\Vol\}_{\pVs} = \{
\Vol_{\pVs_i}, i=1,..,A\}\big)$.
Next, a high resolution volume set $\{\Vol\}_*$ is acquired by applying volume SR
function ($\func$) to each volume in $\{Vol\}_{\pVs}$ (line $3-6$).
Finally, we combine 3D volumes in $\{Vol\}_{*}$ to form a high-resolution 4D
output (line $7$).

The applications of $\texttt{VSR}$ function in  SSR, ASR, and ASSR
are described in Algorithm~\ref{alg:assr}.
In the case of SSR, $\func$ is comprised of PSSR, denoted as $\pssr$, and EVRN
denoted as $\evrn$, (line 3).
$\texttt{VSR}$ is applied to horizontal and vertical volumes separately and then
the output LFs are averaged (line 4).
In the case of ASR, $\func$ is set as PASR, denoted as $\pasr$, followed by EVRN
(line 10).
The super-resolution of horizontal volumes (line 11) and vertical volumes (line
12) will increase the angular resolution to $\Scale_{\Vs}A\times A$ and
$\Scale_{\Vs}A\times \Scale_{\Vs}A$ respectively.
As for ASSR, a similar procedure to ASR is applied, except that PSSR is
previously employed to spatially up-sample the input light field (line 7,8).

\subsection{EPI Volume Refinement Network}
The proposed network bases on global residual learning architecture that is
implemented with a long skip connection and an element-wise addition as
illustrated in Fig.~\ref{fig:evrn}.
Global residual learning allows EVRN to avoid learning complicated
transformation and focus on the reconstruction of high-frequency information
differing between low and high-resolution EPI volume.
3D convolutional kernels are utilized in our network instead of 2D convolutions
since this type of kernel was shown to be effective with 3D EPI volume
data~\cite{Heber2017}.
EVRN is comprised of two parts: attention-based residual learning extracts densely
residual-based features and attention-based multi-path learning reconstructs
high-frequency information.
\subsubsection{Attention-based Residual Learning}
This part consists of a shallow feature extraction layer (SFE) followed by $R$
local residual learning blocks.
Dense connection~\cite{Huang2017} is employed to alleviate gradient vanishing
and improve signal propagation.
With this setup, the feature maps of all preceding layers are combined and used
as an input to the current layer.
Since the accumulated feature size gets bigger after each layer and demands a
high computational effort, we decide to compress the concatenated features by a
feature bottle-neck layer (FBN).
FBN consisting of a $1\Stimes 1\Stimes 1\Stimes C$ convolutional layer followed by a
PReLU activation layer will reduce the dense-feature size to $C$ channels
before inputting to a channel attention-based residual layer (CAR) as in
Fig.~\ref{fig:evrn}.

For the local residual learning block, we follow RCAN~\cite{Zhang2018rcan} to integrate
channel attention~\cite{Woo2018} for adaptively scaling residual features.
As shown in ~\cite{Zhang2018rcan}, this technique improves the reconstruction
quality of high-resolution images.
The architecture of \textit{CAR} is comprised of a well-known feature extraction
combination \textit{conv-prelu-conv} followed by a channel attention weighting
block (CAW).
CAW starts with a global average pooling layer which collapses an input feature
from $W\Stimes A \Stimes H \Stimes C$ to $1\Stimes 1 \Stimes 1 \Stimes C$.
It is then reshaped to $1\Stimes C$ and is followed by a 1D down-sampling convolution
($1\Stimes C_r$) and 1D up-sampling convolution ($1\Stimes C$).
Here $C_r = \frac{C}{r}$ with $r$ is a predefined reduction ratio.
After going through a sigmoid activation layer, the $1\Stimes C$ scaling weight
is reshaped and broadcasted to the form $W\Stimes A \Stimes H \Stimes C$ being
ready for an element-wise multiplication with the input feature.

\subsubsection{Attention-based Multi-Path Learning}
This part is comprised of two separate learning paths targeting spatial and
angular aspects of the feature maps.
Each path includes an FBN, two SFEs, and an attention-based weighting block as in
Fig.~\ref{fig:evrn}.
There are two types of attention-based weighting, one is for spatial feature
dimensions denoted as SAW and the other is for angular feature dimension denoted
as AAW.
FBN is employed to reduce the size of densely connected feature-maps
generated by the residual learning part.
After this block, the feature size is reduced from $(R+1)C$ channels to $C$
channels.
The reduced feature map is fed to an SFE whose output is refined by an
element-wise multiplication with attention weights computed by AAW or SAW.
After the second SFE, the feature maps from the two paths are concatenated and
squeezed to form the final residual feature map.
The high-resolution EPI volume is acquired by adding up the residual information
to the preliminarily up-sampled volume.

SAW consists of a global pooling, a 2D convolution, and a sigmoid activation
layer.
In global pooling layer, Average Pooling ($P_{avg}$) and Max Pooling ($P_{max}$)
are applied to the input feature $F \in \mathbb{R}^{W\Stimes A\Stimes H \Stimes C}$
and results in $F_{avg},\ F_{max} \in \mathbb{R}^{W\Stimes 1\Stimes H\Stimes
  1}$.
These feature maps are then concatenated and reshaped to form the global pooling
feature $F_{pool} \in \mathbb{R}^{W\Stimes H\Stimes 2}$.
We then applied a 2D convolution with kernel size $5\Stimes 5\Stimes 1$ followed
by a sigmoid activation function.
These weighting values are then reshaped and broadcasted to the form $W\Stimes
A\Stimes H \Stimes C$  for an element-wise multiplication.
The following equation summarizes the computation of spatial attention weights.
\begin{equation}
  \label{eq:saw}
  W_{s}(F) = \sigma\bigg(f_{5\Stimes 5\Stimes 1}\Big(\big[P_{avg}(F); P_{max}(F)\big]\Big)\bigg)
\end{equation}

AAW consists of a global pooling, a fully-connected, and a sigmoid activation
layer.
Given an input feature $F \in \mathbb{R}^{W\Stimes H\Stimes A \Stimes C}$, we
follow a similar procedure as of SAW to compute global pooling features
$F_{pool} \in \mathbb{R}^{2A}$.
Here, $P_{avg}$ and $P_{max}$ are applied to spatial and channel dimensions of
the input feature map and the output features are concatenated along the angular
dimension.
We then compute $A$ weighting values by applying a fully connected layer followed
by a sigmoid activation function.
Equation~\ref{eq:aaw} outlines this computation.
\begin{equation}
  \label{eq:aaw}
  W_{a}(F) = \sigma\bigg(D_{2A:A}\Big(\big[P_{avg}(F); P_{max}(F)\big]\Big)\bigg)
\end{equation}

\subsection{Preliminary spatial super-resolution}
\label{sec:pssr}
Given an EPI volume $\Vol_l$ with a resolution of $W\Stimes A\Stimes H$ as an
input, the output of PSSR is an EPI volume $\Vol_h$ with a resolution
of $\Scale_{\Xy}W \Stimes A \Stimes \Scale_{\Xy}H$, where $\Scale_{\Xy}$ is a
spatial scaling factor (i.e., $2,4$).
The procedure of PSSR is as follows. First, 2D images (SAIs) along angular dimension
($\Vol_l(x,i,y), i\!=\!1,2,..,A$) are extracted from the input volume. Secondly, each
image is separately super-resolved to the desired resolution by a SISR method.
Finally, we combine these up-sampled images to form the output volume $\Vol_h$,
see Fig.~\ref{fig:overall}.

In this work, deep learning-based methods (i.e.,
EDSR~\cite{Lim2017}, RCAN~\cite{Zhang2018rcan}) are employed to up-sample SAIs.
With many advanced learning architectures introduced lately and sufficient
training data, deep learning-based approaches easily outperform
optimization-based approaches and provide state-of-the-art performance in the SISR
task.
However, as pointed out in the literature~\cite{Yuan2018, Wang2018, Yeung2019,
  Zhang2019} that SISR alone did not perform well on light field images due to
missing the contribution of external information shared across multiple SAIs.
This such information is well contained in an EPI volume and is exploited in the
proposed EVRN to enhance the output of SISR.
As will be shown later in the experimental result, the proposed approach
substantially improves the reconstruction quality of SISR on both challenging
synthetic and real-world LF data.

\subsection{Preliminary angular super-resolution}
\label{sec:pasr}

In this stage, a novel perspective image is generated for each consecutive pair
of SAIs in the input volume.
This task is done by a novel view synthesis module (NVS) as depicted in
Figure~\ref{fig:overall}.
Given an EPI volume $\Vol_l$ with a resolution of $W\Stimes A \Stimes H$ as an
input, the output of PASR is an EPI volume $\Vol_h$ with a resolution
of $W\Stimes (2A\!-\!1) \Stimes H$.
The angular resolution of the volume is up-sampled by a scaling factor of
$\Scale_{\Vs}=\frac{2A\!-\!1}{A}$.

The synthesis of a novel view can be seen as the interpolation of a novel pixel
along a line on an EPI image, see Fig.~\ref{fig:epi}.
This task is indeed trivial if the slope of this line, also referred to as
disparity value~\cite{Wanner2014}, is known in advance.
Therefore, many previous approaches~\cite{Wanner2014, Kalantari2016, Tran2018}
proposed to firstly estimate disparity maps and then exploit them for
synthesizing novel views.
However, as shown in \cite{Yoon2017, Gul2018}, this explicit estimation of
a disparity map is not necessary since a learning-based approach which directly
inferences novel views can already provide better performance.
In this work, two different approaches of NVS are evaluated, i.e., nvs-cnn and
nvs-mean.
nvs-cnn is the proposed end-to-end CNN learning to synthesize a novel view from
an input image pair.
In nvs-mean, the new view is computed by simply averaging the two input images.
As will be shown in Section~\ref{sec:exp_asr}, this straightforward approach can
provide good results in the case of narrow baseline LF images captured
by plenoptic cameras~\cite{Adelson1992,Ng2005}.

\begin{figure}[t]
  \centering
  \includegraphics[width=0.85\linewidth]{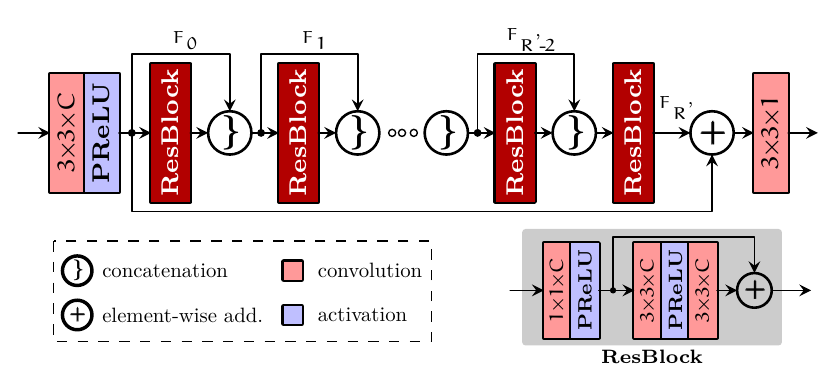}
  \caption{The network architecture of the proposed CNN-based method for PASR}
  \label{fig:nvs_cnn}
\end{figure}

The network architecture of nvs-cnn is presented in Fig.~\ref{fig:nvs_cnn}.
Similar to EVRN, global/local residual learning and dense connection are
employed in the proposed network.
We stack two input images to form an input feature $X\in \mathbb{R}^{W\Stimes
  H\Stimes 2}$. A shallow feature extraction layer consisting of a 2D
convolutional kernel $3\Stimes 3\Stimes C'$ followed by a PReLU activation is applied to $X$.
This results in a feature map $F_0\in \mathbb{R}^{W\Stimes H\Stimes C'}$.
After going through $R'$ layers of residual blocks, we get a residual feature map
$F_{R'}\in\mathbb{R}^{W\Stimes H\Stimes C'}$  which is added to $F_0$ and then
squeezed by a 2D convolution kernel $3\Stimes 3\Stimes 1$ to acquire a novel
perspective image ($I \in \mathbb{R}^{W\Stimes H}$).
For residual block, we use a simple design that consists of a bottleneck layer,
i.e.,  2D convolution kernel ($1\Stimes 1\Stimes C'$) followed by a PReLU, and a
common combination conv-prelu-conv as in Fig.~\ref{fig:nvs_cnn}.
\section{Experimental Results}
\label{sec:exp}
\subsection{Dataset and training}
\begin{table}[t]
  \centering
  \caption{Summary of training and test dataset}
  \begin{tabular}{|l|c|c|c|c|}
    \hline
    Datasets& Type & Angular & Training & Test \\
    \hline
    \hline
    HCI13\cite{Wanner2013} &synthetic & $9\Stimes 9$ & $5$ & $2$ \\
    \hline
    HCI17\cite{Honauer2017} & synthetic & $9\Stimes 9$  & $20$ & $4$ \\
    \hline
    InSyn\cite{Shi2019} & synthetic & $9\Stimes 9$ & $31$ & $8$ \\
    \hline
    StGantry\cite{StanfordGantry} & real-world & $17\Stimes 17$ & $9$ & $3$ \\
    \hline
    StLytro\cite{StanfordLytro} &real-world & $14\Stimes 14$  & $250$ & $53$ \\
    \hline
    InLytro\cite{InriaLytro} &real-world & $15\Stimes 15$ & $28$ & $8$\\
    \hline
    EPFL\cite{Rerabek2016} &real-world  & $15\Stimes 15$ & $81$ & $12$\\
    \hline
    \hline
    Sum & - & - & $424$ & $90$ \\
    \hline
  \end{tabular}
  \label{tab:dataset}
\end{table}

Seven published light field datasets, listed in Table~\ref{tab:dataset}, are
employed for training and testing the proposed approach.
There are three synthetic datasets generated by 3D object models and blender
software~\cite{Wanner2013, Honauer2017, Shi2019}.
The other four datasets are real-world data captured by
Illum cameras~\cite{StanfordLytro, InriaLytro, Rerabek2016} and a gantry
setup~\cite{StanfordGantry}.
While the synthetic scenes have the same angular resolution of $9\Stimes 9$,
the angular resolution of real-world scenes varies from $14\Stimes 14$ to
$17\Stimes 17$.
To have an uniform light field data for training and testing,
we follow the previous work~\cite{Wang2018,Yuan2018,Yeung2019,Zhang2019} to remove the border views and keep only
$9\Stimes 9$ sub-aperture images in the middle.

\subsubsection{Training EVRN}
The LF images in the training set are transformed into YCbCr color space and
only Y channel data is used for training. In inference phase, we apply the trained
network to each color space separately and then convert them back to RGB color
image.
Each LF image is spatially cropped into 4D patch ($P_{H}
\in \mathbb{R}^{48\Stimes 48\Stimes 9\Stimes 9}$) with a stride of 16 pixels.
Plain patches which do not include texture information are ignored.
For each 2D patch with the size of $48\Stimes 48$ from $P_{H}$, we apply
bicubic down-sampling with a scaling factor $\Scale_\Xy$ ($\Scale_\Xy=2, 4$)
to acquire a low spatial resolution patch
$P_{LS}\in \mathbb{R}^{\Scale^{-1}_{\Xy} 48\Stimes \Scale^{-1}_{\Xy} 48\Stimes
  9\Stimes 9}$.
Each 2D patch in $P_{LS}$ is then up-sampled using SISR method to generate a spatial
pre-scaling 4D patch $P_{S}$.
A low angular resolution 4D patch ($P_{LA} \in \mathbb{R}^{
  48\Stimes 48\Stimes 5\Stimes 5} $) is extracted from $P_{H}$ by removing 2D
patches at angular
position $(\rho_i, \tau_i)$, where $(\rho_i \bmod 2) \lor (\tau_i \bmod 2)=1$.
We then up-sampled $P_{LA}$ using the technique discussed in
Section~\ref{sec:pasr} to acquire angular pre-scaling 4D patch $P_{A}$.
An angular-spatial pre-scaling 4D patch $P_{SA}$ is generated by applying the same
procedure to $P_{S}$.
At this point, we have three 4D patch pairs $\big(\{P_{S}\}, \{P_{H}\}\big)$,
$\big(\{P_{A}\},\{P_{H}\}\big)$, and  $\big(\{P_{SA}\},\{P_{H}\} \big)$
to train EVRN for SSR, ASR, and ASSR respectively.
For each 4D patch pair, e.g., $\big(\{P_{S}\},\{P_{H}\}\big)$, a 3D EPI volume
pair  $\big(\{V_{L}\},\{V_{H}\}\big)$  is formed by extracting and combining the
horizontal and vertical EPI volumes from each 4D patch set.

\begin{table}[t]
  \centering
  \caption{Performance analysis of various model configurations.}
  \begin{tabular}{|l|c|c|c|c|}
    \hline
    \multirow{2}{*}{Models}& \multicolumn{2}{c|}{Synthetic} & \multicolumn{2}{c|}{Real-world}  \\
    \cline{2-5}
                           & PSNR & SSIM & PSNR & SSIM \\
    \hline
    \hline
    model 1 & 38.61 & 0.9613 & 38.09 & 0.9581 \\
    \hline
    model 2 & 38.63 & 0.9613 & 38.13 & 0.9588 \\
    \hline
    model 3 & 38.77 & 0.9627 & 38.21 & 0.9592 \\
    \hline
  \end{tabular}
  \label{tab:perf_att}
\end{table}

\begin{table}[t]
  \centering
  \caption{Ablation investiation of attention modules (CAS, AAS, SAW).
    The average PSNRs are computed for dataset EPFL with scaling factor
    $\Stimes 2$ after 200 epochs.}
  \resizebox{\linewidth}{!}{
  \begin{tabular}{|c|c|c|c|c|c|c|c|c|}
    \hline
    Modules & \multicolumn{8}{c|}{Combination of attention modules}\\
    \hline
    \hline
    CAW & \xmark & \cmark & \xmark & \cmark & \xmark & \cmark & \xmark & \cmark\\
    \hline
    SAW & \xmark & \xmark & \cmark & \cmark & \xmark & \xmark & \cmark & \cmark\\
    \hline
    AAW & \xmark & \xmark & \xmark & \xmark & \cmark & \cmark & \cmark & \cmark\\
    \hline
    \hline
    PSNR& 36.92&	36.95&	36.93&	36.95&	36.94&	36.97&	36.94&	36.98\\
    \hline
  \end{tabular}
  }
  \label{tab:abl_att}
\end{table}

Padding is enabled in all 3D convolution layers to reserve the resolution.
The number of residual blocks $R$ and channel size $C$ were set to $7$ and $64$ respectively.
\(\ell_1\) loss function was used since it provides better results for SR in the proposed approach.
The proposed network was implemented in TensorFlow running on a PC with an Nvidia
1080Ti GPU.
As an optimizer, a variation of the Adam optimizer, AdamW~\cite{Loshchilov2019},
was used.
AdamW adds a weight decay as regularization to the Adam optimizer.
The learning rate was initialized to $2\Stimes 10^{-4}$ and decrease by a factor
of 2 after 10 epochs. The \textit{weight decay} was set to $10^{-4}$.
For the initialization of convolution parameters, \textit{Glorot uniform
  initializer} was used.
The bias of the CNN layers and the PReLU parameters were
initialized to zero.
\subsubsection{Training NVS-CNN}
To prepare the training data for NVS-CNN, we first extracted EPI volumes from
4D patches ${P_{H}}$.
For each EPI volume, we use even-index views as ground-truths and the two
neighbor views as inputs. This gives us 4 training pairs for each EPI volume.
We empirically set the number of residual blocks $R'$ and the number of feature
channels $C'$ to 7 and  64 respectively.
Padding is enabled in all convolution layers to preserve the spatial dimension.
For this training, we employ Adam optimizer ($\beta_1=0.9, \beta_2=0.999$).
The learning rate is set to $2\Stimes 10^{-4}$ which is halved after every 10
epochs.
\subsection{Model Analysis}
\label{sec:model_analysis}
To analyze the contribution of attention modules to the performance of the
proposed approach, we conducted an experiment in which three models were tested.
In the first model, all attention modules were removed from the network. The
second model includes only the channel attention module (CAW), and the third model
includes all attention modules.
These models were trained for SSR tasks ($\Scale=2$) and their results are listed
in Table~\ref{tab:perf_att}.
It can be seen from the table that attention modules help to improve the
reconstruction quality.
Without attention modules, \textit{model 1} scores 38.61 dB and 38.09 dB on
average for synthetic and real-world light field data respectively.
These figures slightly increase in \textit{model 2} where the CAW module is
included.
Employing attention modules in both multi-path learning (AAW, SAW) and residual
learning (CAW) provides the highest quality in terms of PSNR and SSIM.

\begin{table*}[h]
  \setlength{\tabcolsep}{0.3em}
  \centering
  \caption{Quantitative comparison of SSR approaches in terms of reconstruction
    quality measured by PSNR/SSIM. The results of two up-scaling factor
    ($\Stimes 2$, $\Stimes 4$) on
    seven public datasets are reported.}
  \resizebox{\textwidth}{!}{
    {\renewcommand{\arraystretch}{1.2}
      \begin{tabular}{|l|c|c|c|c|c|c|c|c|c|}
        \hline
        Approach& Scale&HCI17\cite{Honauer2017}&HCI13\cite{Wanner2013}&InLytro\cite{InriaLytro}&InSyn\cite{Shi2019}&EPFL\cite{Rerabek2016}&StGantry\cite{StanfordGantry}&StLytro\cite{StanfordLytro}&avg\\
        \hline
        \hline
        Bicubic&$\Stimes$2&28.33/0.876&35.02/0.951&30.64/0.900&29.18/0.892&28.30/0.850&33.52/0.965&31.83/0.924&30.97/0.908\\
        \hline
        pcabm\cite{Farrugia2017}&$\Stimes$2&28.94/0.886&35.56/0.949&31.10/0.901&29.69/0.894&28.80/0.853&30.10/0.885&32.05/0.920&30.89/0.898\\
        \hline
        EDSR\cite{Lim2017}&$\Stimes$2&31.51/0.926&39.57/0.973&33.27/0.932&33.70/0.932&30.50/0.890&39.06/0.985&35.62/0.957&34.75/0.942\\
        \hline
        DBPN\cite{Haris2018}&$\Stimes$2&31.93/0.930&39.90/0.974&33.50/0.934&34.27/0.935&30.80/0.893&39.30/0.985&36.00/0.959&35.10/0.944\\
        \hline
        RCAN\cite{Zhang2018rcan}&$\Stimes$2&32.25/0.932&39.91/0.975&33.57/0.934&34.39/0.936&30.90/0.894&39.82/0.986&36.21/0.960&35.29/0.945\\
        \hline
        LFCNN\cite{Yoon2017}&$\Stimes$2&31.67/0.904&38.03/0.963&33.48/0.928&32.12/0.906&32.22/0.926&37.30/0.974&34.91/0.941&34.25/0.935\\
        \hline
        LFnet\cite{Wang2018}&$\Stimes$2&32.31/0.913&39.04/0.968&34.06/0.931&33.13/0.916&32.95/0.931&38.27/0.976&35.53/0.943&35.04/0.940\\
        \hline
        EPI2D\cite{Yuan2018}&$\Stimes$2&33.63/0.929&41.18/0.976&35.15/0.943&35.11/0.930&34.14/0.943&40.36/0.986&37.04/0.956&36.66/0.952\\
        \hline
        SR4D\cite{Yeung2019}&$\Stimes$2&32.49/0.943&39.99/0.977&33.96/0.944&32.29/0.929&30.21/0.889&35.36/0.965&37.53/\textbf{\color{red}0.972}&34.55/0.946\\
        \hline
        resLF\cite{Zhang2019}&$\Stimes$2&35.14/\textbf{\color{red}0.956}&39.84/0.974&35.43/\textbf{\color{red}0.956}&35.27/0.948&34.36/0.954&37.04/0.975&37.67/0.966&36.39/0.961\\
        \hline
        3DVSR-EDSR&$\Stimes$2&\textbf{\color{blue}35.91}/0.954&\textbf{\color{red}43.26}/\textbf{\color{blue}0.984}&\textbf{\color{blue}37.38}/0.955&\textbf{\color{blue}37.01}/\textbf{\color{blue}0.948}&\textbf{\color{blue}36.53}/\textbf{\color{blue}0.955}&\textbf{\color{blue}41.53}/\textbf{\color{blue}0.988}&\textbf{\color{blue}38.74}/0.965&\textbf{\color{blue}38.62}/\textbf{\color{blue}0.964}\\
        \hline
        3DVSR-RCAN&$\Stimes$2&\textbf{\color{red}36.08}/\textbf{\color{blue}0.955}&\textbf{\color{blue}43.20}/\textbf{\color{red}0.984}&\textbf{\color{red}37.46}/\textbf{\color{blue}0.955}&\textbf{\color{red}37.02}/\textbf{\color{red}0.949}&\textbf{\color{red}36.77}/\textbf{\color{red}0.956}&\textbf{\color{red}41.76}/\textbf{\color{red}0.988}&\textbf{\color{red}38.83}/\textbf{\color{blue}0.966}&\textbf{\color{red}38.73}/\textbf{\color{red}0.965}\\
        \hline
        \hline
        Bicubic&$\Stimes$4&24.06/0.694&30.15/0.857&26.66/0.775&24.77/0.765&24.88/0.706&27.37/0.860&26.84/0.790&26.39/0.778\\
        \hline
        pcabm\cite{Farrugia2017}&$\Stimes$4&24.65/0.726&30.77/0.865&27.14/0.792&25.37/0.780&25.37/0.725&26.51/0.790&27.32/0.804&26.73/0.783\\
        \hline
        EDSR\cite{Lim2017}&$\Stimes$4&26.17/0.784&33.83/0.905&28.78/0.830&28.05/0.842&26.67/0.766&30.97/0.932&29.30/0.857&29.11/0.845\\
        \hline
        DBPN\cite{Haris2018}&$\Stimes$4&26.58/0.797&34.19/0.912&29.05/0.835&28.83/0.854&27.01/0.776&31.34/0.936&29.68/0.864&29.52/0.853\\
        \hline
        RCAN\cite{Zhang2018rcan}&$\Stimes$4&26.70/0.801&34.46/0.913&29.07/0.837&28.88/0.856&27.15/0.777&31.66/0.940&29.81/0.867&29.68/0.856\\
        \hline
        LFCNN\cite{Yoon2017}&$\Stimes$4&26.07/0.700&31.47/0.859&28.35/0.796&26.51/0.756&27.22/0.783&28.61/0.861&28.17/0.789&28.06/0.792\\
        \hline
        LFnet\cite{Wang2018}&$\Stimes$4&27.16/0.745&33.33/0.890&29.41/0.829&28.02/0.799&28.43/0.818&30.52/0.900&29.56/0.829&29.49/0.830\\
        \hline
        EPI2D\cite{Yuan2018}&$\Stimes$4&28.15/0.786&35.60/0.914&30.56/0.851&29.66/0.838&29.64/0.847&32.30/0.935&30.70/0.856&30.94/0.861\\
        \hline
        SR4D\cite{Yeung2019}&$\Stimes$4&27.15/0.830&34.08/0.921&29.61/0.864&27.47/0.839&26.84/0.786&28.26/0.854&30.75/\textbf{\color{red}0.896}&29.17/0.856\\
        \hline
        resLF\cite{Zhang2019}&$\Stimes$4&28.83/\textbf{\color{red}0.844}&34.30/0.917&30.63/\textbf{\color{red}0.881}&29.79/\textbf{\color{red}0.875}&29.53/0.869&30.63/0.903&30.80/0.882&30.64/0.882\\
        \hline
        3DVSR-EDSR&$\Stimes$4&\textbf{\color{blue}29.59}/0.833&\textbf{\color{blue}37.07}/\textbf{\color{blue}0.938}&\textbf{\color{blue}31.63}/0.875&\textbf{\color{blue}31.13}/0.870&\textbf{\color{blue}30.85}/\textbf{\color{red}0.872}&\textbf{\color{blue}33.36}/\textbf{\color{blue}0.946}&\textbf{\color{blue}32.07}/0.883&\textbf{\color{blue}32.24}/\textbf{\color{blue}0.888}\\
        \hline
        3DVSR-RCAN&$\Stimes$4&\textbf{\color{red}29.70}/\textbf{\color{blue}0.836}&\textbf{\color{red}37.17}/\textbf{\color{red}0.939}&\textbf{\color{red}31.69}/\textbf{\color{blue}0.876}&\textbf{\color{red}31.23}/\textbf{\color{blue}0.872}&\textbf{\color{red}31.09}/\textbf{\color{blue}0.872}&\textbf{\color{red}33.53}/\textbf{\color{red}0.948}&\textbf{\color{red}32.15}/\textbf{\color{blue}0.885}&\textbf{\color{red}32.37}/\textbf{\color{red}0.890}\\
        \hline
      \end{tabular}
    }
  }
  \vspace{0.2em}

  \footnotesize{\textit{({\color{red}red}: best, {\color{blue}blue}: second best)}}
  \label{tbl:comp_spatial}
\end{table*}
A more comprehensive ablation study of attention modules can be found in
Table~\ref{tab:abl_att}. In this experiment, we investigated the effects of
various combinations of attention modules. The eight networks were trained for
spatial super-resolution application with scaling factor $\Stimes 2$ and have
the same configuration of residual blocks (R=4) and channel size (C=16). RCAN
was employed for the PSSR stage, and the angular size of EPI volume was set to
3. After 200 epochs, we evaluate the performance of trained networks on the EPFL
dataset and report the average PSNR values. From Table~\ref{tab:abl_att}, it can
be seen that the baseline network (without any attention modules) gives the
lowest PSNR value (36.92dB). Furthermore, we observed that channel attention
(CAW) and angular attention (AAW) demonstrate a clear contribution to the
performance of EVRN. While SAW itself provides not much improvement, its
combination with AAW and CAW delivers the best performance (36.98dB).
\subsection{Spatial Super-Resolution}
% ssr x2, x4
\begin{figure*}[t]
  \centering
  \begin{minipage}{1.0\linewidth}
    \includegraphics[width=\linewidth]{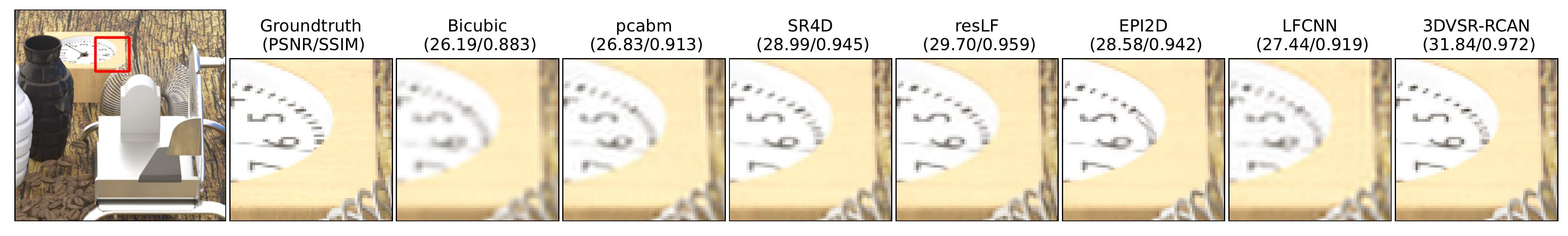}
  \end{minipage}
  \begin{minipage}{1.0\linewidth}
    \includegraphics[width=\linewidth]{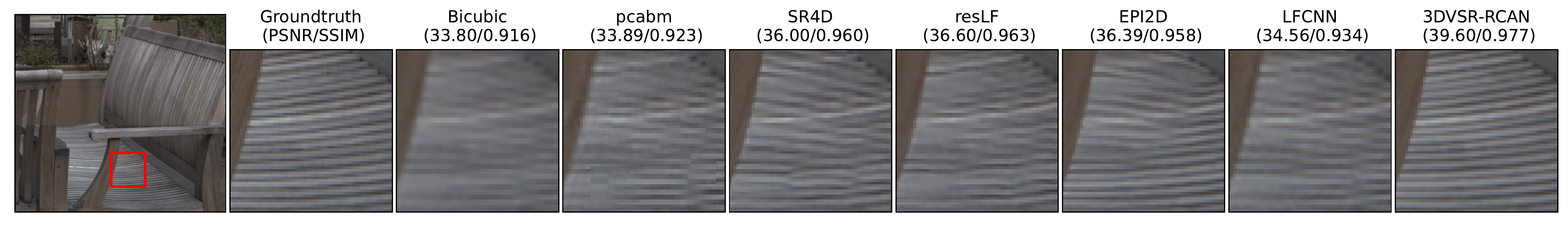}
  \end{minipage}
  \begin{minipage}{1.0\linewidth}
    \includegraphics[width=\linewidth]{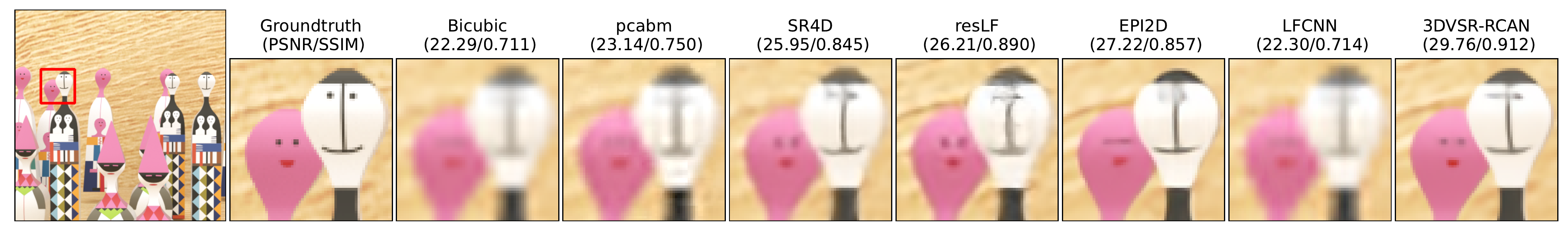}
  \end{minipage}
  \begin{minipage}{1.0\linewidth}
    \includegraphics[width=\linewidth]{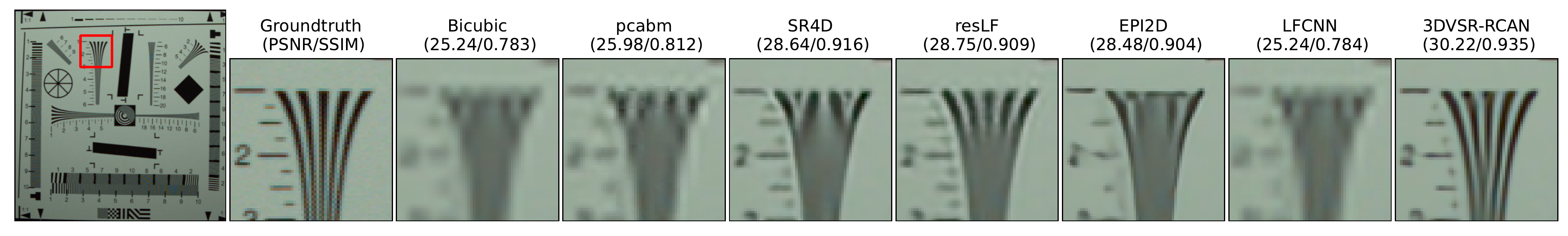}
  \end{minipage}
  \caption{Qualitative comparison of SSR approaches on two scaling factors
    ($\Stimes 2, \Stimes 4$).
    The SAI at $\Vs\!=\!(4,4)$ is visualized together with its
    zoom-in region marked by a red rectangle.
    The first two rows: $\Stimes 2$ results of synthetic scene
    {\it coffee\uds beans\uds vases}~\cite{Shi2019} and real-world scene {\it
      general\uds 55}~\cite{StanfordLytro}.
    The last two rows: $\Stimes 4$ results of synthetic scene {\it smiling\uds
      crowd}~\cite{Shi2019} and real-world scene
    {\it ISO\uds Chart}~\cite{Rerabek2016}.
  }
  \label{fig:ssr_x}
\end{figure*}
\begin{figure}[t]
  \centering
  \begin{minipage}{\linewidth}
    \includegraphics[width=\textwidth]{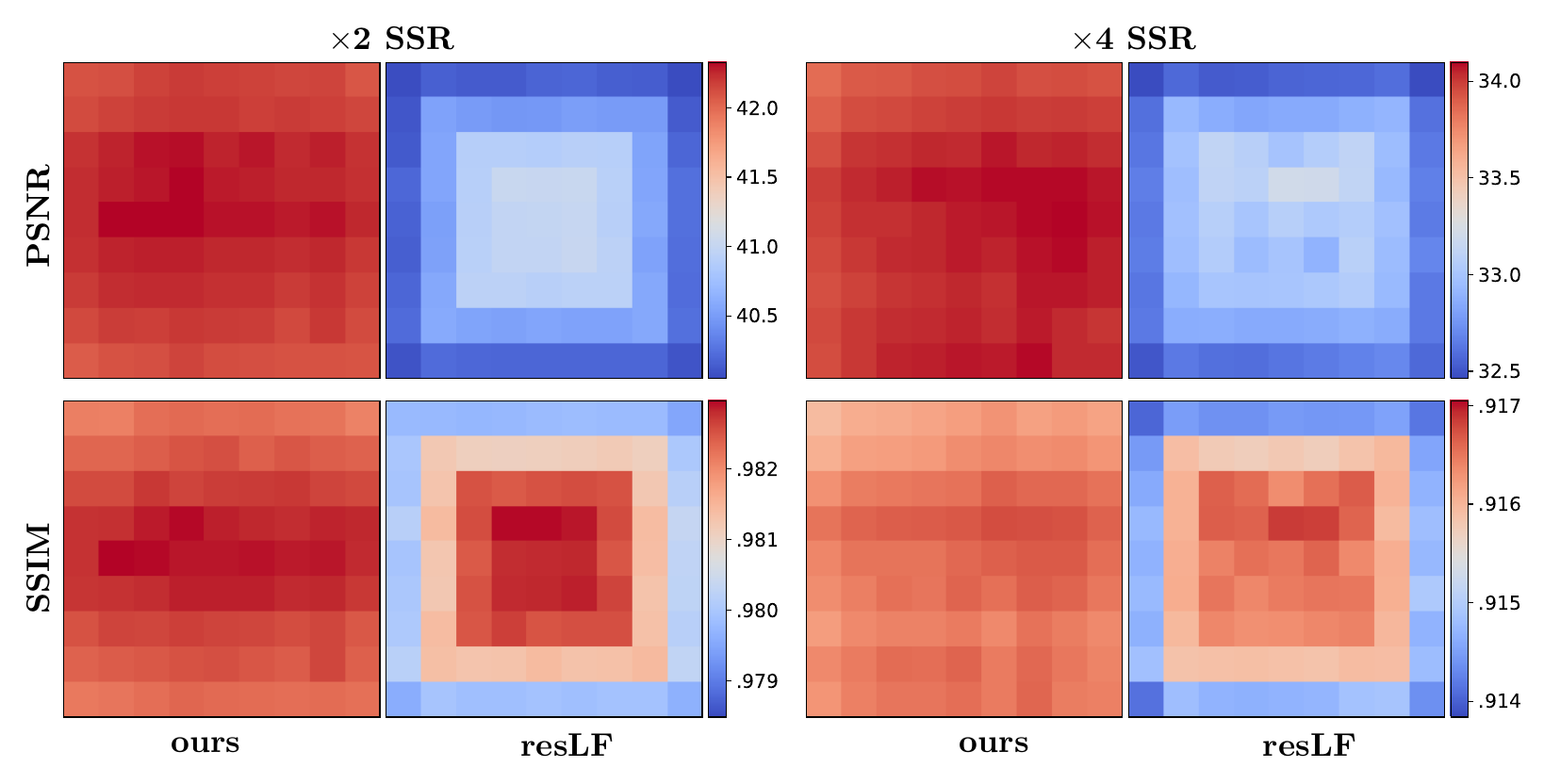}
  \end{minipage}
  \caption{Visualization of PSNR and SSIM values on each SAI of light field scene
    \textit{two\uds vases}~\cite{Shi2019}.
    Compared to ResLF~\cite{Zhang2019}, the proposed approach achieve a better
    reconstruction quality while maintaining a consistent performance on all
    perspective images.
  }
  \label{fig:ssr_grid}
\end{figure}
In this section, the evaluation results of 3DVSR applied for SSR problem are
discussed.
We employed EDSR~\cite{Lim2017} and RCAN~\cite{Zhang2018rcan} for preliminary
spatial super-resolution and tested against two scaling factors $\Scale_\Xy\!=\!2$
and $\Scale_\Xy\!=\!4$.
Nine state-of-the-art approaches are selected for quantitative and qualitative
comparisons. Among them, there are three SISR approaches (EDSR~\cite{Lim2017},
DBPN~\cite{Haris2018}, RCAN~\cite{Zhang2018rcan}) and six approaches provided
for 4D LF (pcabm~\cite{Farrugia2017}, LFnet~\cite{Wang2018}, LFCNN~\cite{Yoon2017},
EPI2D~\cite{Yuan2018}, SR4D~\cite{Yeung2019}, resLF~\cite{Zhang2019}).
The result of bicubic interpolation is presented as a baseline result.
All methods, except LFCNN, EPI2D, and LFnet, were tested using their released
codes and pre-trained models.
For EPI2D and LFnet, since the authors did not publish their source codes, we
followed their papers to implement the models.
Although the source code of LFCNN is available, its pre-trained model is not
provided. Therefore, we retrained LFCNN, as well as EPI2D and LFnet, using our
training dataset.

Table~\ref{tbl:comp_spatial} lists quantitative results for $\Stimes 2$ and
$\Stimes 4$ in terms of PSNR and SSIM metrics.
For each scene, the quality metrics are calculated as an average of $7\Stimes 7$
SAIs in the middle. These values are then averaged over the test dataset.
The two configurations of PSSR using EDSR and RCAN are denoted as 3DVSR-EDSR and
3DVSR-RCAN respectively.
It can be seen from the table that the proposed approach scores the best PSNR and
SSIM value for both $\times 2$ and $\times 4$ problems on average.
Compared to SISR approaches EDSR and RCAN, 3DVSR respectively improves the
reconstruction quality by 3.87dB and 3.63dB for $\Stimes 2$ and 3.13dB and
2.67dB for $\Stimes 4$.
This improvement pays a tribute to the proposed enhancement network (EVRN) which exploits
EPI volume structure to correct the high-frequency information from the output
of SISR.
The advantage of using EPI volume is also evident when compared to the 2D EPI-based
approach~\cite{Yuan2018}. Using a similar SISR technique (i.e. EDSR), the
proposed approach scores 1.96 dB and 1.4dB better on average for $\Stimes 2$ and
$\Stimes 4$ respectively.
\begin{figure*}[t]
  \centering
  \begin{minipage}{\linewidth} % row
    \centering
    \includegraphics[width=\textwidth]{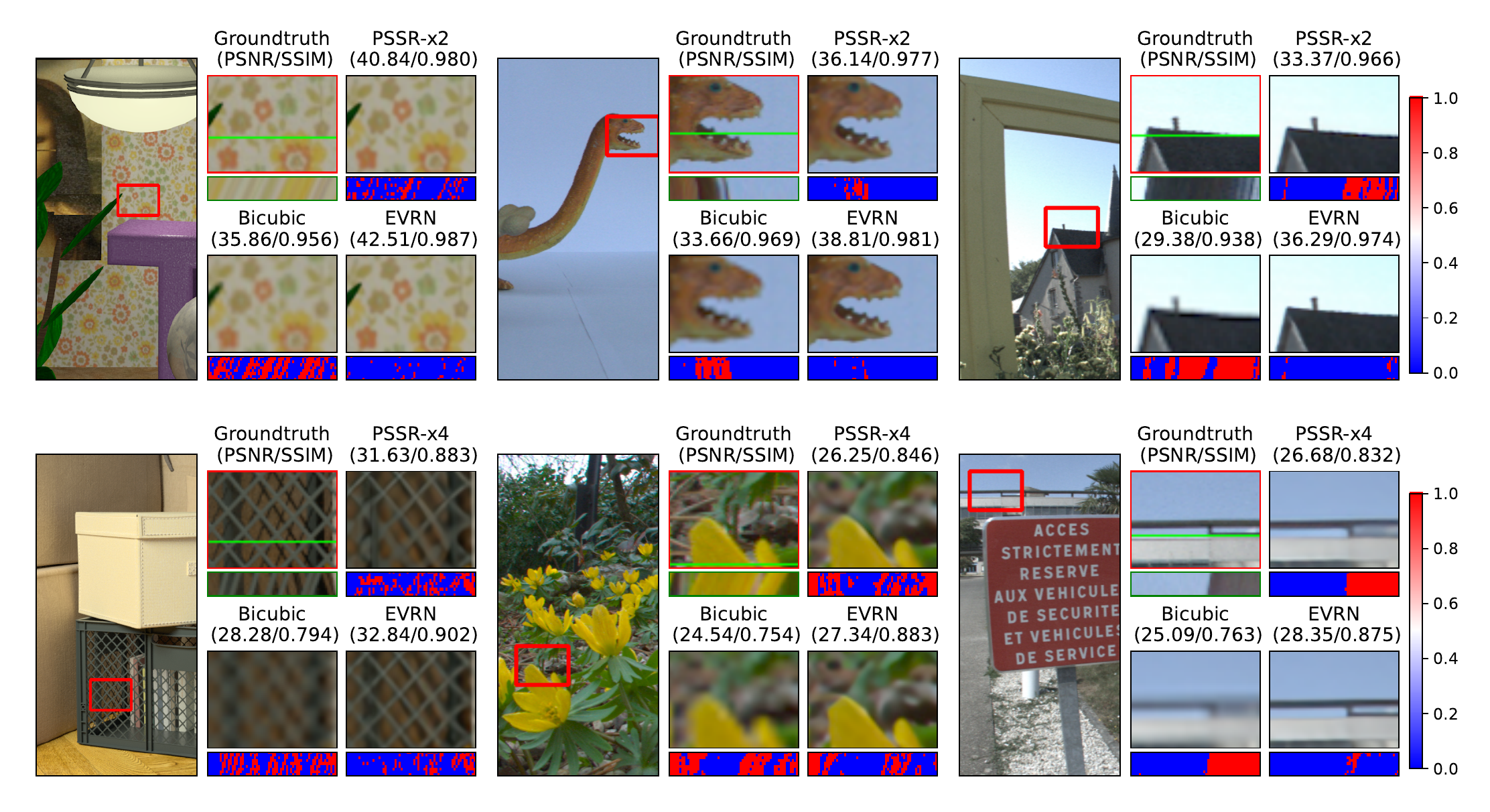}
  \end{minipage}
  \caption{
    Comparison of two stages in reconstruction of spatially high resolution LF.
    The SAI at $\Vs\!=\!(4,4)$ is visualized together with its
    zoom-in region marked by a red rectangle. For each approach, an EPI at
    horizontal line marked in green is extracted and compared to the EPI of Groundtruth.
    The first row, from left to right: $\Stimes 2$ results of synthetic scene
    {\it mona}~\cite{Wanner2013} and real-world scenes {\it Ankylosaurus\uds\&\uds
      Diplodocus\uds 1}, {\it Framed}~\cite{Rerabek2016}.
    The second row, from left to right: $\Stimes 4$ results of synthetic scene
    {\it boxes}~\cite{Honauer2017} and real-world scenes {\it Flowers}, {\it Sign}~\cite{Rerabek2016}.
  }
  \label{fig:ssr_2s}
\end{figure*}

Qualitative comparisons of six light field SSR approaches are shown in
Fig.~\ref{fig:ssr_x}.
It can be observed from the figures that the proposed approach shows superior
performance in visual effects for both synthetic and real-world scenes.
For example, only 3DVSR can reconstruct a correct pattern of wooden sticks in
\textit{general\uds 55} and clear detail of two faces in \textit{Smilling\uds
  crowd}.
In pcabm~\cite{Farrugia2017}, the low-resolution light field image is
divided into sets of patches that are then super-resolved separately.
Although the overlapping of patches helps to smooth the output images, the
inconsistencies in reconstruction quality between patches are unavoidable and lead
to blocky artifacts (i.e. \textit{ISO\uds Chart}, \textit{Smilling\uds
  crowd}).
In LFCNN~\cite{Yoon2017}, each SAI is super-resoluted independently using a SISR
approach~\cite{Dong2014} whose CNN is quite trivial and shallow.
LFCNN's results are, therefore, over-smoothed in all the test scenes.
SR4D~\cite{Yeung2019} and resLF~\cite{Zhang2019} include multiple SAIs in their
super-resolution process.
This allows them to exploit external information to reconstruct high-resolution
images.
However, these approaches still fail to reconstruct high quality images in
challenging light field scenes with noisy and high-frequency pattern (i.e.
\textit{general\uds 55}) or highly degraded content (i.e.,
$\Stimes 4$ down-sampled: \textit{ISO\uds Chart} and \textit{Smilling\uds crowd}).
From the figure, their results are ambiguous and with obvious artifacts.
Compared to EPI2D, we follow a similar approach in which the output of SISR is
enhanced by reinforcing EPI structures of light field images.
However, instead of relying on a narrow and feature-limited EPI as in EPI2D,
we proposed to refine an EPI volume that allows us to exploit global information
across multiple EPIs.
Our results, therefore, show significantly better image qualities where the
texture is recovered correctly and with more high-frequency details.

To investigate the reconstruction quality concerning different perspectives, we
conducted an experiment in which the scene \textit{two\uds vases} is used as
input to perform both $\times 2$ and $\times 4$ SSR. For each view, the PSNR and
SSIM values are computed, and the results are visualized in
Fig.~\ref{fig:ssr_grid}. Here, we compared our results with the results of
state-of-the-art approach ResLF~\cite{Zhang2019}. resLF employs a star-like
structure of SAIs as input to super-resolute a single SAI lying in the center.
This strategy is indeed beneficial to reconstruct high-resolution images. As in
Table~\ref{tbl:comp_spatial}, resLF surpasses EPI2D and SR4D for most of the
test dataset. However, the disadvantage of resLF's approach is unequal treatment
of SAI from different perspectives. For example, the SAI closed to the border
has fewer input images than the SAI closed to the center. Therefore,
reconstruction qualities of non-central views are relatively lower. In contrast,
the proposed approach jointly uses EPI information and spatial information to
reconstruct an EPI volume and thus achieves much higher output qualities with
more stable distribution across all perspectives.

Fig.~\ref{fig:ssr_2s} presents a quantitative and qualitative comparison of EPI
volumes reconstructed after the first stage (PSSR) and the second stage (EVRN).
To better justify the improvement in angular dimension, we extract EPI images
for each output and compare them to the EPI of the ground truth. From the
figure, it is evident that the EVRN substantially improves the quality of
reconstructed volume in both spatial and angular dimensions.

\begin{table*}
  \setlength{\tabcolsep}{0.3em}
  \centering
  \caption{Quantitative comparison of ASR approaches on 7 light field datasets}
  \resizebox{\textwidth}{!}{
    {\renewcommand{\arraystretch}{1.2}
      \begin{tabular}{|l|c|c|c|c|c|c|c|c|}
        \hline
        Approach&HCI17\cite{Honauer2017}&HCI13\cite{Wanner2013}&InLytro\cite{InriaLytro}&InSyn\cite{Shi2019}&EPFL\cite{Rerabek2016}&StGantry\cite{StanfordGantry}&StLytro\cite{StanfordLytro}&avg\\
        \hline
        \hline
        nvs-mean&29.23/0.878&42.25/0.987&40.35/0.979&27.73/0.828&37.81/0.975&25.53/0.728&40.35/\textbf{\color{blue}0.981}&34.75/0.908\\
        \hline
        nvs-cnn&37.02/0.971&44.15/0.992&40.49/0.975&38.76/0.979&37.89/0.974&30.02/0.847&40.75/0.980&38.44/0.960\\
        \hline
        vsyn\cite{Kalantari2016}&23.34/0.665&29.41/0.764&32.03/0.899&21.40/0.613&27.76/0.794&19.35/0.542&30.34/0.887&26.23/0.738\\
        \hline
        LFSR\cite{Gul2018}&-/-&-/-&39.91/0.980&-/-&37.63/0.979&22.24/0.688&39.57/0.977&34.84/0.906\\
        \hline
        LFCNN\cite{Yoon2017}&30.97/0.883&38.78/0.967&35.36/0.948&29.09/0.834&34.43/0.949&26.01/0.739&36.41/0.957&33.01/0.897\\
        \hline
        Wang18\cite{Wang2018a}&31.66/0.893&41.74/0.984&37.31/0.967&29.36/0.831&36.18/0.969&27.04/0.741&38.86/0.971&34.59/0.908\\
        \hline
        Wu19\cite{Wu2019}&29.93/0.917&30.76/0.832&33.73/0.922&28.67/0.871&29.48/0.843&20.80/0.631&32.45/0.937&29.40/0.850\\
        \hline
        Wu19a\cite{Wu2019a}&33.19/0.936&43.28/0.985&39.90/0.969&32.12/0.899&37.68/0.966&27.51/0.746&40.03/0.973&36.24/0.925\\
        \hline
        3DVSR-mean&\textbf{\color{red}40.12}/\textbf{\color{blue}0.979}&\textbf{\color{red}47.07}/\textbf{\color{red}0.994}&\textbf{\color{red}44.22}/\textbf{\color{red}0.985}&\textbf{\color{blue}40.32}/\textbf{\color{blue}0.979}&\textbf{\color{red}43.64}/\textbf{\color{red}0.987}&\textbf{\color{red}32.30}/\textbf{\color{blue}0.854}&\textbf{\color{red}43.41}/\textbf{\color{red}0.985}&\textbf{\color{red}41.58}/\textbf{\color{blue}0.966}\\
        \hline
        3DVSR-cnn&\textbf{\color{blue}40.00}/\textbf{\color{red}0.981}&\textbf{\color{blue}45.15}/\textbf{\color{blue}0.994}&\textbf{\color{blue}43.04}/\textbf{\color{blue}0.982}&\textbf{\color{red}40.90}/\textbf{\color{red}0.982}&\textbf{\color{blue}42.30}/\textbf{\color{blue}0.986}&\textbf{\color{blue}32.11}/\textbf{\color{red}0.861}&\textbf{\color{blue}42.20}/0.980&\textbf{\color{blue}40.81}/\textbf{\color{red}0.967}\\
        \hline
      \end{tabular}
    }
  }
  \vspace{0.2em}

  \footnotesize{\textit{({\color{red}red}: best, {\color{blue}blue}: second best)}}
  \label{tbl:comp_angular}
\end{table*}

\begin{figure*}[t]
  \centering
  \begin{minipage}{1.0\linewidth}
    \includegraphics[width=\linewidth]{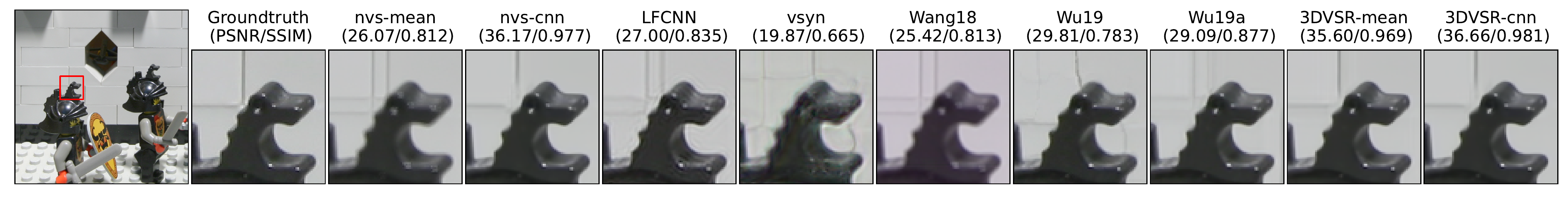}
  \end{minipage}
  \begin{minipage}{1.0\linewidth}
    \includegraphics[width=\linewidth]{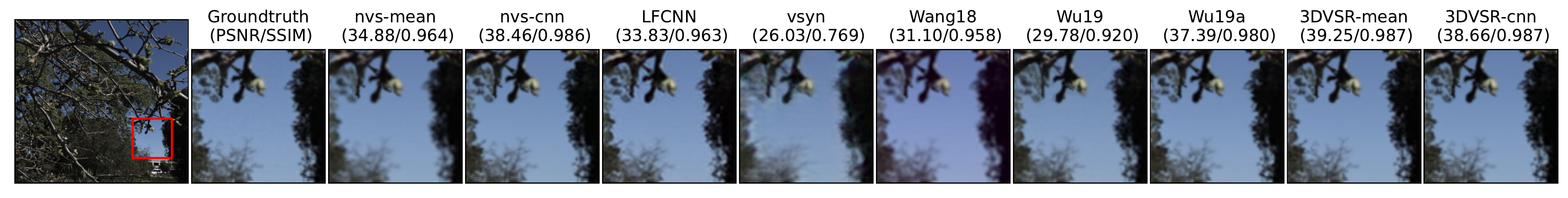}
  \end{minipage}
  \begin{minipage}{1.0\linewidth}
    \includegraphics[width=\linewidth]{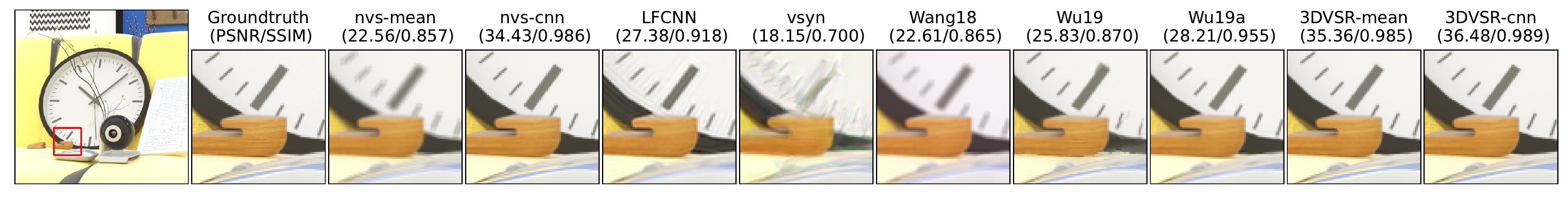}
  \end{minipage}
  \begin{minipage}{1.0\linewidth}
    \includegraphics[width=\linewidth]{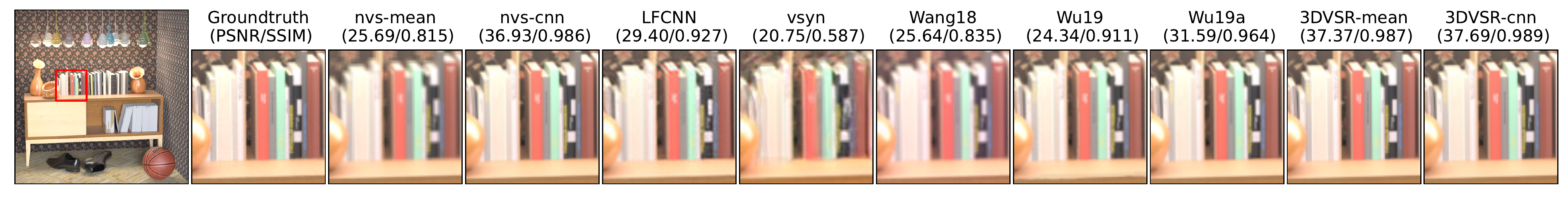}
  \end{minipage}
  \caption{Qualitative comparison of ASR approaches on real-world and synthetic
    light field scenes. The SAI at $\Vs\!=\!(3,3)$ is visualized together with its
    zoom-in region marked by a red rectangle.
    The first two rows: real-world scenes \textit{lego}~\cite{StanfordGantry}
    and \textit{flowers\uds plants\uds 36}~\cite{StanfordLytro}.
    The last two rows:
    synthetic scenes \textit{big\uds clock}~\cite{Shi2019} and
    \textit{sideboard}~\cite{Honauer2017}.
  }
  \label{fig:asr_1}
\end{figure*}

\subsection{Angular Super-Resolution}
\label{sec:exp_asr}

This section discusses the evaluation of the proposed approach for the angular
super-resolution of light field images. The seven datasets listed in
Table~\ref{tab:dataset}, are employed in this evaluation. For each scene in the
test set, the angular resolution is down-sampled from $9\Stimes 9$ to $5\Stimes
5$ while the spatial resolution remains intact. With these low-resolution LFs as
inputs, we then reconstruct the original size LFs follow the procedures
discussed in Section~\ref{sec:method}. This means that 56 missed perspective
images are reconstructed from 25 input images. In the PASR stage, we tested two
approaches; one generates novel views by averaging, and the other using an
end-to-end CNN. The results of these two PASR approaches and the final results
after applying the refinement network are reported. We compare our approach with
six previous approaches (vsyn~\cite{Kalantari2016}, LFCNN~\cite{Yoon2017},
LFSR~\cite{Gul2018},  Wang18~\cite{Wang2018a}, Wu19~\cite{Wu2019}, and
Wu19a~\cite{Wu2019a}).

Table~\ref{tbl:comp_angular} lists quantitative results of ASR approaches
running on the seven public datasets. We employed PSNR and SSIM as quality
metrics that are computed for newly generated SAIs and are averaged over all
scenes in each dataset. {\it nvs-mean} and {\it nvs-cnn} denotes the two PASR
approaches. {\it nvs-mean} computes a novel perspective image by averaging two
neighboring images, while {\it nvs-cnn} inferences a novel image using a
residual CNN as depicted in Fig.~\ref{fig:nvs_cnn}. The outputs of {\it
  nvs-mean} and {\it nvs-cnn} enhanced by EVRN are denoted as {\it 3DVSR-mean}
and {\it 3DVSR-cnn} respectively.
LFSR~\cite{Gul2018} has a strict requirement of supported angular resolution.
This approach employs a fully connected network in its output layer that always
produces an angular-resolution of $14\Stimes 14$. For this reason, only
real-world datasets~\cite{StanfordLytro, InriaLytro, Rerabek2016,StanfordGantry}
are tested with this approach.
From Table~\ref{tbl:comp_angular}, it can be seen that the proposed approach
provides the highest reconstruction quality. 3DVSR improves PSNR and SSIM values
by a large margin as compared to the previous approaches (i.e., a minimum of 3dB
improvement in all test datasets).
In narrow-baseline light field images captured by a plenoptic
camera~\cite{InriaLytro, Rerabek2016, StanfordLytro}, the difference between two
neighboring views is almost invisible due to their sub-pixel displacement values
(e.g., less than 0.5 pixel). In this case, a straightforward approach such as
nvs-mean can score very well, and the benefit of employing CNN in the PASR stage
is limited, i.e., the improvement of nvs-cnn over nvs-mean is smaller than
0.4dB. However, for the other test datasets, nvs-cnn presents a clear
improvement compared to nvs-mean (i.e., 1.9dB to 7.8dB). By exploiting the EPI
volume structure to refine the results of the PASR stage, 3DVSR achieves
significant improvements over nvs-mean and nvs-cnn by an average of 6.8 dB and
2.4 dB, respectively. It is interesting to see that the performance of
3DVSR-mean is comparable to 3DVSR-cnn with a slightly better PSNR value (i.e.,
0.7 dB). This demonstrates the superior performance of EVRN in reconstructing
novel perspective images.

\begin{figure*}
  \centering
  \begin{minipage}{\linewidth}
    \centering
    \includegraphics[width=\textwidth]{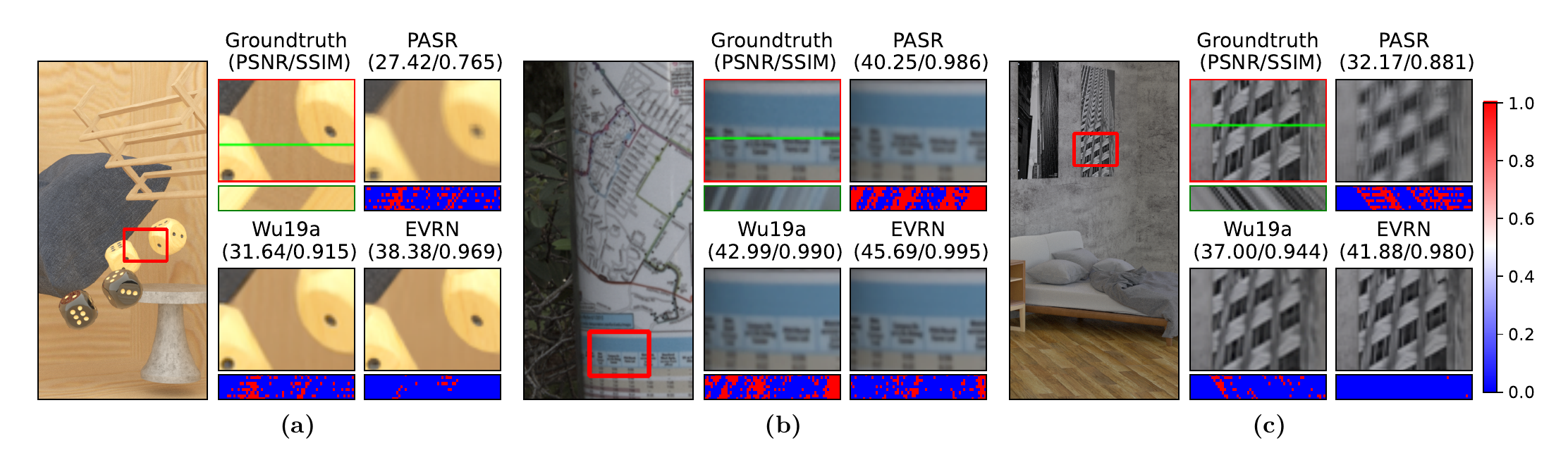}
  \end{minipage}
  \caption{
    Comparison of two stages in reconstruction of angularly high resolution LF.
    The SAI at $\Vs\!=\!(3,3)$ is visualized together with its
    zoom-in region marked by a red rectangle. For each approach, an EPI at
    horizontal line marked in green is extracted and compared to the EPI of Groundtruth.
    From left to right: synthetic scene {\it Flying\uds dice\uds dense}~\cite{Shi2019},
    real-word scene {\it general\uds 29}~\cite{StanfordLytro},
    and synthetic scene {\it bedroom}~\cite{Honauer2017}.
  }
  \label{fig:asr_2s}
\end{figure*}

\begin{figure*}[ht]
  \centering
  \begin{minipage}{.9\linewidth}
    \includegraphics[width=\linewidth]{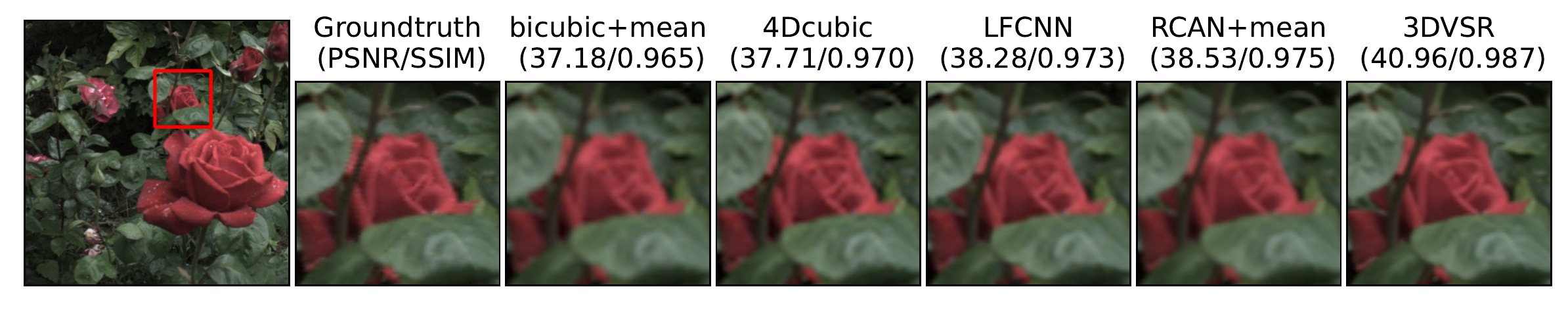}
  \end{minipage}
  \begin{minipage}{.9\linewidth}
    \includegraphics[width=\linewidth]{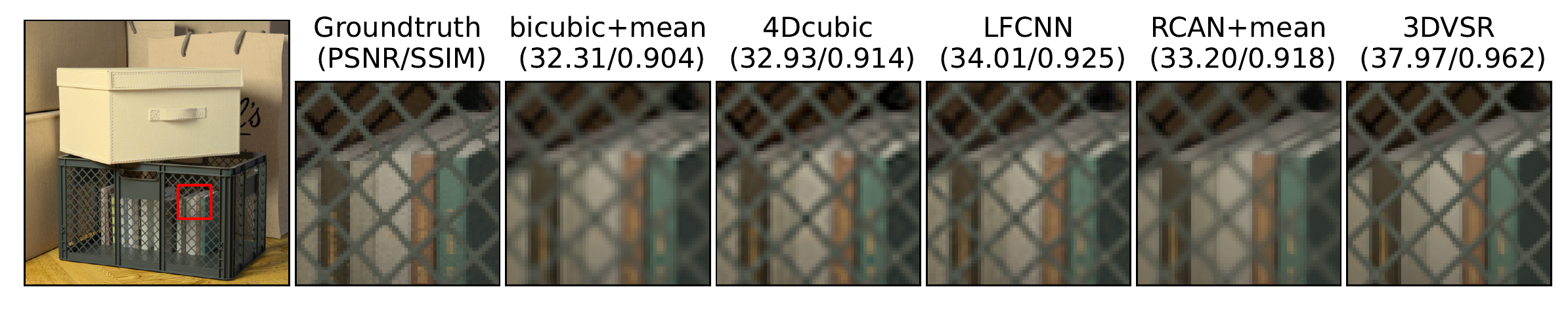}
  \end{minipage}
  \caption{
    Angular-spatial super-resolution results of different approaches.
    The SAI at $\Vs\!=\!(3,3)$ is visualized together with
    its zoom-in region marked by a red rectangle.
    top row: real-world scene \textit{Rose} ~\cite{InriaLytro};
    bottom row: synthetic scene \textit{boxes}~\cite{Wanner2013};
  }
  \label{fig:assr_1}
\end{figure*}

Fig.~\ref{fig:asr_1} shows the visual comparisons of evaluated ASR approaches on
synthetic and real-world light field scenes. vsyn~\cite{Kalantari2016} consists
of two CNNs; one predicts auxiliary disparity maps, and the other synthesizes
novel views. It assumes that the disparity values of input light fields fall
within a range that is quantized by a fixed step size. Based on the quantized
disparity values, a cost volume is computed and is used as an input to the first
CNN. The output disparity maps from the first CNN are used to pre-compute novel
views that are then refined by the second CNN. The quality of the novel views
highly depends on the accuracy of estimated disparity maps which are in turn
depends on the cost volume and assumed disparity range. From the figure, it can
be seen that vsync does not generalize well and performs poorly on light field
scenes with a large disparity range (i.e., lego, big\uds clock, sideboard).
Although the visual quality of the plenoptic scene (i.e., flowers\uds plants\uds
36), for which vsyn was trained, is much better, it still suffers from the
over-smoothed effect. Wang18~\cite{Wang2018a} employ a 3D CNN to recover the
high-frequency detail of a stack of SAIs. Since their network architecture is
relatively shallow and straightforward, it performs not so well and leaves a
visible blurry effect on the synthesized images. Taking advantage of EPI
structure, Wu19~\cite{Wu2019} and Wu19a~\cite{Wu2019a} can reconstruct images
with more detail than Wang18. Both LFCNN and our cnn-based PASR approach
(nvs-cnn) follow a similar strategy in which novel perspective images are
synthesized by using their surrounding neighbor images. However, as opposed to
the simple architecture of LFCNN, which only consists of convolution layers and
activation layers, nvs-cnn employed many effective deep learning structures
(i.e., global/local residual learning, dense connection). nvs-cnn, therefore,
outperforms LFCNN with much better visual quality. Compared to nvs-cnn, the
novel perspective images generated by nvs-mean are more ambiguous with
over-smoothed regions and artifacts as the result of the averaging method.
However, after being refined by EVRN the visual qualities of these images are
significantly enhanced (i.e., 3DVSR-mean) and are comparable to the enhanced
version of nvs-cnn (i.e., 3DVSR-cnn).

Fig.~\ref{fig:asr_2s} compares the angular super-resolution results of
Wu19a~\cite{Wu2019a} and the proposed approaches after the first stage (PASR)
and after the second stage (EVRN). In this experiment, nvs-mean is employed as a
preliminary angular super-resolution approach, and the output volume of PASR is
fed to EVRN for the final enhancement. Compared to our PASR, Wu19a provides a
better reconstruction quality with sharper content and less angular error.
However, the refined volume of EVRN is by far better than the output of Wu19a.
As a result, we achieve a minimum of 3.3dB improvement in PSNR and less error in
EPIs.
\subsection{Angular-Spatial Super-resolution}
\label{sec:exp_assr}
In the ASSR problem, the resolution of 4D light field images is super-resoluted
angularly and spatially. In other words, it consists of a super-resolution of
each given SAI and a synthesis of novel perspective images which have the same
higher resolution. As discussed in Section~\ref{sec:method}, the proposed
approach handle ASSR in two stages. The first stage consists of PSSR followed by
PASR to generate a 4D light field with the desired resolution. EVRN then
enhances this preliminary up-sampled light field in the second stage to acquire
the final output. To evaluate the proposed approach, we conducted an experiment
in which the spatial scaling factor was set to $\Scale_\Xy=2$ and a similar ASR
configuration as in Section~\ref{sec:exp_assr} was applied.

The spatial-angular super-resolution results are shown in Fig.~\ref{fig:assr_1}.
\textit{bicubic+mean} denotes a baseline approach that employs bicubic
interpolation for SSR and averages neighbor views to generate novel perspective
images. \textit{4Dcubic} denotes an approach in which cubic interpolation is
applied on two spatial and two angular dimensions. \textit{RCAN+mean} denotes
the result of our preliminary stage which uses
\textit{RCAN}~\cite{Zhang2018rcan} for PSSR and \textit{nvs-mean} for PASR. The
result after an enhancement using EVRN is denoted as \textit{3DVSR}. Compared to
the baseline approach and 4Dcubic, LFCNN provides a small improvement in
reconstruction quality. Although LFCNN achieves an increase of about 1dB, its
improvement in visual quality is negligible. The results of LFCNN are still
ambiguous and lack high-frequency information. A similar performance can be seen
in the results of \textit{RCAN+mean} which are mostly over-smoothed due to the
effect of averaging views. Compared to these approaches, 3DVSR produces a
significant improvement in PSNR value (i.e., a minimum of 2.4 dB and 3.9 dB in
scenes \textit{Rose} and \textit{boxes} respectively) and an obvious enhancement
in visual quality.

\section{Conclusion}
\label{sec:conclusion}
This paper presents an angular-spatial light field super-resolution approach
being able to reconstruct high-quality 4D light fields. Based on EPI volume
structure, a 3D projected version of a 4D light field, we proposed a 2-stage
framework that effectively addresses various problems in light field
super-resolution, i.e., ASR, SSR, and ASSR. While the earlier stage provides
flexible options to up-sample the input volume to the desired resolution, the
later stage, which consists of an EPI volume-based enhancement CNN,
substantially improves the reconstruction quality of the high-resolution EPI
volume. The proposed enhancement network built on 3D convolutional operations
and efficient deep learning structures, i.e., global/local residual learning,
dense connection, multi-path learning, and attention-based scaling, effectively
combines angular and spatial information from the 3D EPI volume structure to
reconstruct high-frequency details. An extensive evaluation on 90 challenging
synthetic and real-world light field scenes from 7 published datasets shows that
the proposed approach outperforms state-of-the-art methods to a large extend for
both spatial and angular super-resolution problems, i.e., an average PSNR
improvement of more than 2.0 dB, 1.4 dB, and 3.14 dB in SSR $\Stimes 2$, SSR
$\Stimes 4$, and ASR respectively.
The reconstructed 4D light field demonstrates a balanced performance
distribution across all perspective images and presents superior visual quality
compared to the previous works.

\end{document}